\documentclass[journal]{IEEEtran}
\usepackage{amsmath,amsfonts}
\usepackage{algorithm}
\usepackage{algpseudocode}
\usepackage{array}
\usepackage{subcaption}
\usepackage{xspace}

\newcommand{\name}{{\sc PACE }}
\newcommand{\mame}{{\sc PACE}}

\usepackage{textcomp}
\usepackage{stfloats}
\usepackage{url}
\usepackage{verbatim}
\usepackage{graphicx}
\usepackage{cite}
\usepackage{hyperref}
\usepackage{amssymb}
\usepackage{colortbl}
\usepackage{multirow}
\usepackage{array}
\usepackage{booktabs}
\usepackage[dvipsnames]{xcolor}
\definecolor{LightCyan}{rgb}{0.88,1,1}
\usepackage{enumitem} 
\usepackage{csquotes}
\usepackage{cleveref}
\Crefname{figure}{Fig.}{Figs.}
\crefname{figure}{Fig.}{Figs.}
\usepackage{marvosym}

\begin{document}

\title{[CLS] and [EOS] Tokens are not Enough \\
for CLIP-based Class-Incremental Learning}

\author{Hao Sun, Tao Hu, Shixiong Kai, Siyuan Xu, Wenqian Zhao, Mingxuan Yuan, Da-Wei Zhou%
\thanks{This work was supported in part by the Basic Research Program of Jiangsu (BK20251251), NSFC (62506160), JSTJ2025-147, Fundamental and Interdisciplinary Disciplines Breakthrough Plan of the Ministry of Education of China (No. JYB2025XDXM118), the 111 Center (No. B26023), and the Collaborative Innovation Center of Novel Software Technology and Industrialization. (Corresponding author: Da-Wei Zhou)

Hao Sun, Tao Hu and Da-Wei Zhou are with the School of Artificial Intelligence and the State Key Laboratory for Novel Software Technology, Nanjing University, Nanjing, China (e-mail: sunhao@lamda.nju.edu.cn, hut@lamda.nju.edu.cn, zhoudw@lamda.nju.edu.cn).

Shixiong Kai, Siyuan Xu, Wenqian Zhao, and Mingxuan Yuan are with Huawei Noah’s Ark Lab (e-mail: 
kaishixiong@huawei.com, xusiyuan520@huawei.com, zhao.wenqian@huawei.com, Yuan.Mingxuan@huawei.com)
}}
      
\markboth{Journal of \LaTeX\ Class Files,~Vol.~14, No.~8, August~2021}%
{Shell \MakeLowercase{\textit{et al.}}: A Sample Article Using IEEEtran.cls for IEEE Journals}

\maketitle
\begin{abstract}
Class-Incremental Learning (CIL) aims to empower models to continuously acquire new knowledge while mitigating catastrophic forgetting. With the advent of vision-language models like CLIP, leveraging their cross-modal alignment capabilities to address CIL challenges has emerged as a dominant paradigm. Existing methods leverage CLIP's global representations (\textit{i.e.}, the visual [CLS] token and the textual [EOS] token) for cross-modal alignment. While achieving effective classification, they overlook the rich patch-level visual features and token-level textual semantics inherent in CLIP. For example, the [CLS] token aggregates local information to capture the general concept of ``cat", while the patch-level details retain specific localized features essential for recognition, such as the tail, fur, and face. We propose \name (PAtch-token Cross-modal Enhancement) to uncover the long-neglected local representations within CLIP. First, based on the global representations like [CLS] token and [EOS] token, we identify the top-$\textit{k}$ most crucial visual patches  and textual tokens. Subsequently, to better utilize the extracted key local features, we design a cross-modal fusion module. This module leverages local textual semantics to refine visual representations, while augmenting textual semantics with local visual features. As the fusion module is shared across all tasks, directly fine-tuning it on new tasks would interfere with the knowledge learned from previous tasks, leading to catastrophic forgetting. To alleviate this, we employ a projection-based update strategy to effectively preserve the critical parameter space of previous tasks during the adaptation to new tasks. Extensive experiments demonstrate that \name significantly outperforms state-of-the-art methods across multiple CIL benchmarks. Code is available at \href{https://github.com/LAMDA-CL/PACE}{https://github.com/LAMDA-CL/PACE}.
\end{abstract}
\begin{IEEEkeywords}
Catastrophic Forgetting, Class-Incremental Learning, CLIP, Continual Learning.
\end{IEEEkeywords}
\section{Introduction}\label{sec1}
  \IEEEPARstart{I}{n} recent years, the rapid advancement of deep learning~\cite{he2016deep,shi2025feature,xiao2023clip} has exerted a profound influence across numerous fields, demonstrating significant advantages and potential. However, real-world scenarios are inherently dynamic. Data typically emerges as continuous, evolving data streams~\cite{rebuffi2017icarl,he2025long}, posing substantial challenges for traditional deep learning models. To address this challenge, Class-Incremental Learning (CIL)~\cite{de2021continual,qi2025adaptive,wang2025rethinking,roy2022class} has been proposed as a new research paradigm, enabling models to incrementally absorb new knowledge. A primary obstacle for CIL is catastrophic forgetting~\cite{yang2022continual},~\cite{ma2023forgetting}, in which the model tends to fit current data while forgetting previously acquired knowledge. With the rapid development of Pre-trained Models (PTMs)~\cite{peng2023sgva},~\cite{guo2025prompt}, such as Vision Transformers~\cite{dosovitskiy2020image} and CLIP~\cite{radford2021learning}, leveraging their inherent generalization capabilities has become a dominant trend~\cite{jia2021scaling,hu2025hierarchical}. Consequently, current CIL~\cite{wang2023attriclip} research has shifted from traditional training from scratch with randomly initialized weights to utilizing existing PTMs, which require training only a small number of additional parameters~\cite{yi2024leveraging,zhou2023learning,zhang2025visual}.

CLIP~\cite{radford2021learning}, as a pioneering pre-trained vision-language model~\cite{yuan2021florence}, maps images and texts into a shared embedding space via contrastive learning, effectively achieving cross-modal alignment with powerful zero-shot generalization and semantic representation capabilities. Specifically, the image encoder of CLIP prepends a learnable [CLS] token to the input sequence of image patches to aggregate global visual features, while the text encoder utilizes the [EOS] token at the end of the sequence to summarize the overall textual semantics. Although both encoders simultaneously output a sequence of patch-tokens containing rich spatial details~\cite{wu2024llmclip},~\cite{dai2025propvg} and a sequence of word-tokens carrying local semantics during processing, CLIP's cross-modal matching mechanism relies solely on computing the cosine similarity between the global embeddings. While this design achieves efficient global alignment, it inherently disregards the local feature sequences that harbor specific details during the final matching stage~\cite{menon2022visual}. For instance, when distinguishing between a ``\textbf{sparrow}" and a ``\textbf{robin}", the image explicitly contains a clear ``beak" patch. However, since the matching process relies solely on holistic semantic features, these critical local details cannot be effectively utilized during mapping.

Recent studies~\cite{yu2024memory},~\cite{huang2025mind} in CIL have focused on Parameter-Efficient Fine-Tuning (PEFT) strategies, such as prompt learning~\cite{zhou2022conditional},~\cite{wu2024controlmllm} and adapter tuning~\cite{yu2024boosting},~\cite{gao2024clip} to mitigate catastrophic forgetting. Despite their effectiveness in balancing stability and plasticity, these methods largely inherit the global alignment paradigm of the original CLIP. Specifically, existing works focus on optimizing learnable prompts or refining global [CLS] and [EOS] representations. Consequently, they also overlook the discriminative potential of both visual patch-tokens and textual tokens, leaving the rich local spatial features and local semantic contexts, which are crucial for class-incremental learning.

Leveraging the local information carried by patch-tokens offers a promising approach to enhancing CLIP’s discriminative capabilities in class-incremental learning. However, two major challenges remain: \textbf{1)} Acquiring and effectively utilizing local information. Although both the visual and text encoders of CLIP output rich local features, this information is not effectively utilized by the existing matching mechanism. Furthermore, since image patches contain substantial redundant noise, and the text includes non-informative stop words. So we need to select discriminative local features and design a novel matching paradigm that enables the effective utilization of patch-tokens to jointly contribute to the classification decision. \textbf{2)} Mitigating forgetting when utilizing local knowledge. As class-incremental learning involves continuous updates to the data stream, the model risks forgetting previously acquired knowledge after each update. Therefore, it is crucial to retain learned local knowledge effectively and mitigate catastrophic forgetting throughout the updating process.

To address these challenges, we propose \name (PAtch-token Cross-modal Enhancement) for CLIP-based CIL. First, we address the critical issue of the availability of local features. We uncover the long-neglected local representations within CLIP, extracting the crucial top-$k$ patch-level visual features and token-level textual semantics essential for discrimination, and map them into their respective modal shared spaces. Subsequently, to fully leverage these key local elements, we design a cross-modal fusion module. Token-level textual semantics are utilized to refine the global image representation, while patch-level visual features are simultaneously injected into the text embeddings to enhance their semantic specificity. Finally, the global shared weights of the cross-modal fusion module are highly susceptible to destructive gradient updates from new tasks. To prevent catastrophic forgetting during updates, we adopt a projection-based parameter update strategy. This mechanism effectively preserves the critical parameter space associated with previously acquired knowledge. Our method achieves state-of-the-art performance, demonstrating significant improvements across all nine benchmark datasets.

\section{Related Work}
\subsection{Class-Incremental Learning}
Class-Incremental Learning aims to continuously acquire knowledge from new tasks while preventing catastrophic forgetting~\cite{cao2025class},\cite{wang2024relational}. Typical CIL methods can be broadly categorized into several distinct groups. Replay-based methods~\cite{chaudhry2018riemannian},~\cite{chaudhry2018efficient} mitigate forgetting by maintaining a small memory buffer to store representative exemplars of old classes, which are then replayed during the training of new tasks~\cite{luo2023class},~\cite{zheng2024multi}. Model rectification-based methods rectify the inductive biases within the model, such as classifier weights~\cite{zhao2020maintaining},~\cite{wu2019large} and biased logits~\cite{yu2020semantic}. Dynamic network-based methods~\cite{wang2022foster},~\cite{zhou2022model} dynamically adjust or expand the network structure during the incremental process to better accommodate new classes, typically involving neuron expansion~\cite{xu2018reinforced}, backbone expansion~\cite{wang2022beef}, and prompt expansion~\cite{DBLP:conf/cvpr/LiuSS21}. Knowledge distillation-based methods~\cite{rebuffi2017icarl},~\cite{hinton2015distilling} leverage distillation techniques to transfer knowledge from the old model to the current one, constraining the model to balance between learning new tasks while retaining old ones~\cite{douillard2020podnet},~\cite{li2016learning}. Regularization-based methods impose constraints on critical parameters, minimize their deviation, and thus resist forgetting~\cite{aljundi2019task},~\cite{kirkpatrick2017overcoming}.
\subsection{Pre-Trained Model-Based CIL }
The advent of powerful Pre-trained Models (PTMs), such as ViT~\cite{dosovitskiy2020image} and CLIP~\cite{radford2021learning}, has precipitated a paradigm shift in CIL, moving from training models from scratch with randomly initialized weights to harnessing pre-trained foundations~\cite{qi2025adaptive}. To preserve the pre-trained knowledge, a common strategy is to freeze the PTM backbone and add lightweight, learnable modules for adaptation~\cite{smith2023coda},~\cite{rebuffi2017learning}. For example, Prompt-based methods (\textit{e.g.}, L2P~\cite{wang2022learning}, DualPrompt~\cite{wang2022dualprompt}) design a visual prompt pool~\cite{jia2022visual} that utilizes learnable prompts to encode task-specific information into the input sequence of the frozen backbone. Recent work has further optimized the dynamic adaptability of prompt generation by introducing mechanisms such as attention modules~\cite{smith2023coda} and generative networks~\cite{jung2023generating}. Adapter-based methods~\cite{yu2024boosting},~\cite{Fukuda_2025_CVPR}, on the other hand, achieve adaptation by inserting lightweight modules into the frozen network layers. Building upon this foundation, a prominent trend has emerged that leverages the robust cross-modal alignment capabilities of Vision-Language Models (VLMs), such as CLIP. Specifically, PROOF~\cite{zhou2023learning} augments CLIP's representational capacity by appending novel projection layers. RAPF~\cite{huang2024class} introduces a strategy for adaptive representation adjustment and decomposed parameter fusion. By decomposing parameters into shared and task-specific components via weighted fusion, this method effectively mitigates catastrophic forgetting. Furthermore, CLG-CBM~\cite{Yu_2025_CVPR} incorporates language-guided concept bottleneck models. To maintain semantic consistency across tasks, these models align visual features into a human-interpretable concept space. {MG-CLIP~\cite{huang2025mind} introduces modality gap preservation and compensation to mitigate forgetting and improve adaptation to new tasks, while HERMAN~\cite{xie2025herman} proposes hierarchical representation matching for CLIP-based CIL. However, existing CLIP-based CIL methods still primarily rely on global representations for incremental recognition, leaving the patch-level visual features and token-level textual semantics within CLIP insufficiently explored. This paper aims to explore the long-neglected local representations within CLIP for improving incremental recognition in CLIP-based CIL.}

\section{Preliminaries}
In this section, we introduce the background information of CIL and provide an overview of the CLIP model. Finally, we discuss the typical approaches for CLIP-based CIL.
\subsection{Class-Incremental Learning}
Class-Incremental Learning (CIL) aims to construct a unified classifier capable of recognizing all seen classes from a sequential data stream~\cite{rebuffi2017icarl},~\cite{li2023cross}. Consider a training sequence consisting of $B$ tasks, denoted as $\left\{\mathcal{D}^{1}, \mathcal{D}^{2}, \cdots, \mathcal{D}^{B}\right\}$, where each task $\mathcal{D}^{b}=\left\{\left(\mathbf{x}_{i}, y_{i}\right)\right\}_{i=1}^{n_b}$ 
contains $n_b$ examples. Here,  $\mathbf{x}_{i}\in  \mathbb{R}^D$ is an instance of  class $y_i\in Y_b$, and $Y_b$ denotes the label space corresponding to task $b$. The label spaces from different tasks are mutually disjoint (\textit{i.e.}, $Y_b  \cap Y_{b^\prime} = \varnothing$ for $b\neq b^\prime$). Under the exemplar-free setting ~\cite{wang2022dualprompt},~\cite{zhou2022learning}, the model is prohibited from accessing any historical samples from task $\mathcal{D}^{1:b-1}$ when adapting to the new task $\mathcal{D}^{b}$. The objective of CIL is to construct a unified classifier for all seen classes $\mathcal{Y}_b=Y_1 \cup \cdots \cup Y_b$. Formally, we aim to learn a model $f(\mathbf{x}): X\rightarrow\mathcal{Y}_b$ that minimizes the expected risk:
\begin{equation} \label{eq:totalrisk} 
 	f^*=\underset{f \in \mathcal{H}}{\operatorname{argmin}} \; \mathbb{E}_{(\mathbf{x}, y) \sim \mathcal{D}_{t}^1\cup\cdots\mathcal{D}_{t}^b} \mathbb{I}\left(y \neq f(\mathbf{x})\right) \, ,
 \end{equation}
  where $\mathcal{H}$ represents the hypothesis space, $\mathbb{I}(\cdot)$ is the indicator function, and $\mathcal{D}_{t}^b$ denotes the data distribution of the $b$-th task. 

In line with~\cite{zhou2023learning},~\cite{huang2024class}, we assume the availability of a pre-trained model CLIP ~\cite{radford2021learning} to serve as the initialization for $f(\mathbf{x})$. CLIP consists of an image encoder $g_i(\cdot)$ and a text encoder $g_t(\cdot)$, which project images and texts into a shared $d$-dimensional embedding space.\\
\textbf{Image Encoder}. Given an input image $\mathbf{x} \in \mathbb{R}^{D}$, the image encoder $g_i(\cdot)$ (\textit{e.g.}, Vision Transformer) first splits it into $M$ fixed-size patches. These patches are then flattened and linearly projected into patch embeddings. A learnable class token ([CLS] token) is prepended to the sequence, and positional encodings are added to form the input sequence for the Transformer layers. The sequence subsequently passes through stacked Transformer layers. In each layer, utilizing the multi-head self-attention mechanism~\cite{dosovitskiy2020image}, [CLS] token actively attends to all $M$ patch embeddings in the sequence, thereby aggregating global semantic information layer by layer. Simultaneously, each patch embedding attends to other tokens, allowing local features to acquire global context from the entire image, thereby enriching their own representations. After passing through the transformer layers, we obtain a sequence of feature embeddings:
\begin{equation} \label{eq:2} 
 	\textbf{V} = [\textbf{v}_\text{cls},\textbf{v}_1,\cdots,\textbf{v}_M] \in \mathbb{R}^{(M+1)\times d},
 \end{equation}
where $g_i(\mathbf{x}) = \textbf{v}_\text{cls}\in \mathbb{R}^d$ represents the global visual representation used for alignment, and $\{\textbf{v}_i\}_{i=1}^M$ denotes the sequence of local patch tokens containing rich local feature information.\\
\textbf{Text Encoder}. Similarly, for a given class $y_i\in \mathcal{Y}_b$, we utilize a templated prompt $\mathbf{c}_{i} $(\textit{e.g.}, ``a photo of a [CLASS]"). The text encoder $g_t(\cdot)$ tokenizes the prompt, prepends a start token ([SOS] token) and appends an end token ([EOS] token), processing it into a sequence of embeddings:
\begin{equation} \label{eq:3} 
\textbf{T}^i =  [\textbf{t}_{\text{sos}}^i,\textbf{t}_1^i,\cdots,\textbf{t}_{N}^i,\textbf{t}_{\text{eos}}^i] \in \mathbb{R}^{(N+2)\times d},
 \end{equation}
where $g_t(\mathbf{c}_{i}) =\textbf{t}_\text{eos}^i\in \mathbb{R}^d$ serves as the global semantic representation used for alignment, while the intermediate token sequence $\{\textbf{t}_j^i\}_{j=1}^N$ captures rich local semantic contexts.

During inference, CLIP predicts the category of the input image $\mathbf{x}$ by computing the similarity between its global visual feature $\mathbf{v}_\text{cls}$  and the global text embeddings $\mathbf{t}_\text{eos}$ of all seen classes in $\mathcal{Y}_b$. The probability for class $y_i$ is computed as:
 \begin{align} \label{eq:clip_pred}
	f_{y_i}(\mathbf{x}, \mathbf{c}_{i}) & =\frac{\exp \left(\cos \left(g_i(\mathbf{x}), g_t(\mathbf{t}_i)  \right) / \tau\right)}{\sum_{j=1}^{|\mathcal{Y}_b|} \exp \left(\cos \left(g_i(\mathbf{x}), g_t(\mathbf{t}_j) \right) / \tau\right)} \\\
	\notag
	&= \frac{\exp \left(\cos \left(\textbf{v}_\text{cls}, \textbf{t}_\text{eos}^{i}  \right) / \tau\right)}{\sum_{j=1}^{|\mathcal{Y}_b|} \exp \left(\cos \left(\textbf{v}_\text{cls}, \textbf{t}_\text{eos}^{j} \right) / \tau\right)} ,
\end{align}
where $\cos(\cdot,\cdot)$ denotes cosine similarity and $\tau$ is temperature. $g_i(\mathbf{x})$ and $g_t(\mathbf{t}_i)$ are visual and textual features, respectively. Actually, this matching process is performed using the [CLS] token $\textbf{v}_\text{cls}$ in Eq.~\ref{eq:2} and the [EOS] token $\textbf{t}_\text{eos}^{i}$ in Eq.~\ref{eq:3}.
\subsection{Baselines in Class-Incremental Learning}
To overcome forgetting, there are two typical solutions for
CIL with pre-trained models using PEFT, \textit{e.g.}, prompt learning and adapter tuning.\\
\textbf{Prompt Learning.} Instead of fine-tuning the backbone, these methods insert learnable tokens into the input sequence to instruct the frozen encoders. Methods like ~\cite{wang2022learning},~\cite{wang2022dualprompt} design a prompt pool $\mathcal{P}$ to select instance-specific prompts, with which the adapted visual embedding via the frozen image encoder $\bar{g_i}(\cdot)$ is represented as:
\begin{equation}\label{eq:5} \tilde{\textbf{v}}_\text{cls} = \bar{g_i}(\mathbf{x}, \mathcal{P}) \in \mathbb{R}^d.
\end{equation} By leveraging Eq.~\ref{eq:5}, methods employ the prompt-based embedding  $\tilde{\textbf{v}}_\text{cls}$	 to replace the conventional visual embedding. This strategy enables the model to efficiently encode downstream task-specific information into the visual prompts, thereby achieving effective adaptation to new tasks while mitigating catastrophic forgetting.\\
\textbf{Adapter Tuning.} 
In contrast to prompt learning, several methods~\cite{Fukuda_2025_CVPR},~\cite{cui2025cmoa} propose adding lightweight adapter modules for subsequent tasks. Specifically, the visual encoders are frozen, and adapter modules $\textbf{M}_v$ are added after them to learn task-specific knowledge. The adapted feature is represented as:
\begin{align}
\label{eq:77}
    \tilde{\textbf{v}}_\text{cls} = \textbf{M}_v(\bar{g}_i(\mathbf{x})),
\end{align}
where $\bar{g}_i(\mathbf{x}) = \textbf{v}_\text{cls}$ is  the frozen visual embedding. The adapted feature $\tilde{\textbf{v}}_\text{cls}$ is then substituted for ${\textbf{v}}_\text{cls}$ in Eq.~\ref{eq:77} to compute the classification probabilities. This strategy allows the model to learn downstream knowledge through lightweight adapter modules while mitigating catastrophic forgetting.\\
 \textbf{Discussions.} 
These methods essentially retain the original CLIP matching paradigm, consequently neglecting the patch-level visual features $\{\textbf{v}_i\}_{i=1}^M$ and token-level local textual semantics $\{\textbf{t}_j^i\}_{j=1}^N$ generated in Eq.~\ref{eq:2} and Eq.~\ref{eq:3}. Hence, it is crucial to investigate methods for leveraging local features to improve the model's performance.
 \begin{figure*}[t]
	\vspace{-3mm}
	\begin{center}
		{\includegraphics[width=1\linewidth, height=0.6\linewidth]{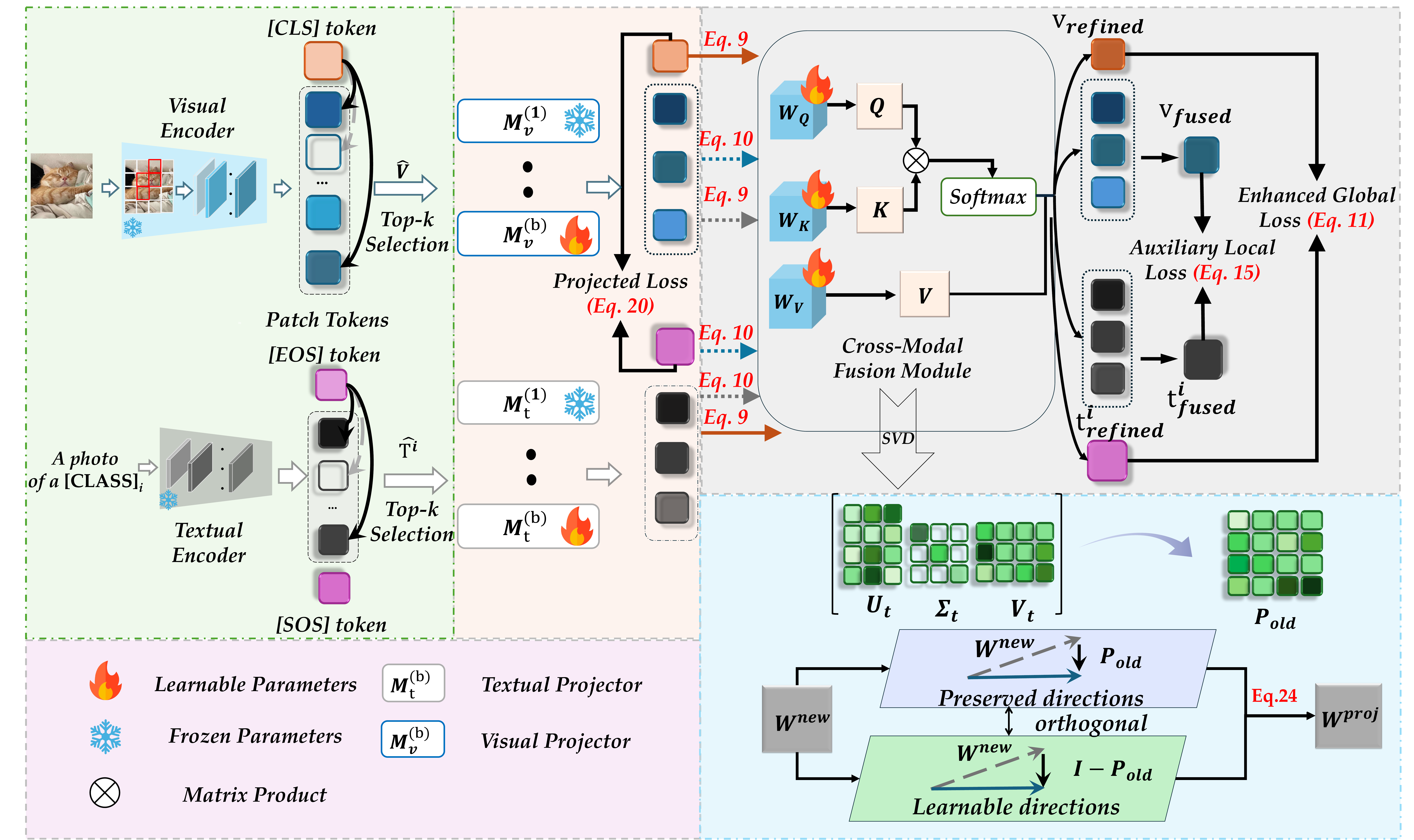}}
	\end{center}
	\vspace{-1mm}
	\caption{  Illustration of \mame. \textbf{Left}: Patch-token information
selection, where top-$k$ discriminative patch-level visual features and token-level textual semantics are selected. \textbf{Top Right}: Cross-modal fusion, which utilizes a cross-attention fusion mechanism. Here, one modality's global representation serves as the query, while the other's local feature sequence serves as the key and value, thereby enhancing the cross-modal semantic representation of the global features. \textbf{Bottom Right}: Local knowledge preservation, which utilizes subspace projection to constrain shared weight updates, thus mitigating catastrophic forgetting.
	}
	\vspace{-3mm}
	\label{figure:teaser}
\end{figure*}
\section{\name: PAtch-token Cross-modal Enhancement}
To fully leverage the rich local information contained within CLIP, enabling the model to adapt to downstream tasks and prevent forgetting, we propose \mame. In Sec.~\ref{sec:4.1}, we first extract the top-$k$ most discriminative local features from the visual patch-level and textual token-level sequences. We map the features of each modality to an independent shared space to enhance adaptability in downstream tasks. Subsequently, in Sec.~\ref{sec:4.2}, we design a cross-modal fusion module that uses global features as queries to aggregate cross-modal local features, thereby explicitly injecting local knowledge into the global representations. To mitigate catastrophic forgetting, we leverage a subspace projection-based parameter update strategy to protect the critical weight subspaces associated with local feature utilization in Sec.~\ref{sec:4.3}. Lastly, Sec.~\ref{sec:4.4} provides a summary of \name and outlines the detailed guidelines for training and inference.

\subsection{ Patch-Token Information Selection}\label{sec:4.1}
Given an input image $\mathbf{x}$, we employ the frozen CLIP visual encoder $\bar{g}_i(\cdot)$ as the feature extractor. Distinct from standard CLIP paradigms that solely extract the  [CLS] token for classification, we explicitly extract the complete sequence of feature embeddings $\textbf{V} = [\textbf{v}_\text{cls},\textbf{v}_1,\cdots,\textbf{v}_M]$ from the final transformer layer using Eq.~\ref{eq:2}, where $\textbf{v}_\text{cls}$ represents the global feature and the subsequence $\{\textbf{v}_i\}_{i=1}^M$ preserves the patch-level spatial details. Symmetrically, for the textual branch, we utilize the fixed text prompt template (\textit{e.g.}, ``A photo of a [CLASS]") as input. We extract the full token sequence $\textbf{T}^i =  [\textbf{t}_{sos}^i,\textbf{t}_1^i,\cdots,\textbf{t}_{N}^i,\textbf{t}_\text{eos}^i]$ from the frozen CLIP text encoder's last layer using Eq.~\ref{eq:3}, where $\textbf{t}_{eos}^i$ aggregates global semantics, while $\{\textbf{t}_j^i\}_{j=1}^N$ captures token-level features. 
  
However, the vast background regions in images (\textit{e.g.}, sky, grass) and non-informative functional stop words in texts introduce noise that significantly hampers the utilization of local features for classification. To address this, for the visual branch, we use the global feature $\textbf{v}_\text{cls}$ as the criterion for assessing the discriminative power of local tokens, because the global feature fully interacts with all local patches via the self-attention mechanism within the transformer layers. We measure the discriminative power of each patch $\textbf{v}_i$ by computing the  normalized affinity with the global feature $\textbf{v}_\text{cls}$. The discriminative power score $s_i^{vis}$ is formulated as:
\begin{align}
    s_i^{vis} &= \text{Softmax}\left(\frac{\text{cos}\left(\textbf{v}_\text{cls},\textbf{v}_i\right)}{\tau}\right)
   \\
	\notag
	&= \frac{\exp \left(\cos \left(\textbf{v}_\text{cls}, \textbf{v}_i  \right) / \tau\right)}{\sum_{j=1}^{M} \exp \left(\cos \left(\textbf{v}_\text{cls}, \textbf{v}_j \right) / \tau\right)} .\
\end{align}
\\Based on the discriminative
power distribution $\{s_i^{vis}\}_{i=1}^M$, for each image $\mathbf{x}$, we select the top-$k$ largest scores and their corresponding patches to construct the discriminative local patch set $\hat{\mathcal{V}}$:
\begin{align}\label{eq:9}
    \hat{\mathcal{V}} = \{\textbf{v}_i| s_i^{vis} \in \text{Top}_k(\{s_j^{vis}\}_{j=1}^M)\} \in \mathbb{R}^{k\times d},
\end{align}
where $\operatorname{Top}_k(\{s_j^{vis}\}_{j=1}^M)$ indicates selecting the top-$k$ largest values in $\{s_j^{vis}\}_{j=1}^M$. This top-$k$ selection is recomputed at every forward pass for each input image, since the visual token sequence varies with the current sample. Symmetrically, for each class $i$, we perform the same operation to extract the top-$k$ tokens $\hat{\mathcal{T}}^i$ that are most tightly coupled with the global context $\textbf{t}_\text{eos}^i$. Since the prompt template and text encoder are fixed, the selected top-$k$ tokens for each class can be precomputed once and reused across iterations. \\

\subsection{Cross-Modal Enhancement}\label{sec:4.2}
After extracting the most discriminative local features in $ \hat{\mathcal{V}}$ and $\hat{\mathcal{T}}^i$ 
 in Sec.~\ref{sec:4.1}, these rich local details remain isolated from the global decision-making embeddings (\textit{i.e.}, $\textbf{v}_\text{cls}$
  and  $\textbf{t}_\text{eos}^i$) used for final classification, making it difficult to fully realize the discriminative potential of the local information. To address this issue, we introduce the cross-modal fusion module to leverage local features ($\hat{\mathcal{V}}$ or $\hat{\mathcal{T}}^i$) to enhance the global representation ($\textbf{v}_\text{cls}$ or $\textbf{t}_\text{eos}^i$) of the other modality. Specifically, the global representation of one modality acts as the query, aiming to enhance the cross-modal semantic representation capability of that global feature by selectively attending to the local feature sequence of the other modality (serving as the key and value). This mechanism establishes two semantic fusion branches, which we detail below. 

\noindent \textbf{Textual-Token to Visual.} In this branch, we aim to inject local textual attributes into the global visual representation. We adopt a \textbf{single-head attention mechanism} for this cross-modal fusion. The global visual feature $\textbf{v}_\text{cls}$ acts as the \textbf{query}, actively retrieving information from the local text tokens $\hat{\mathcal{T}}^i$ (serving as the \textbf{key} and \textbf{value}). Formally, we introduce three learnable projection matrices:  $\mathbf{W}_Q^v$ projects the global visual token into the query space, and $\mathbf{W}_K^v, 
 \mathbf{W}_V^v$ project the local textual tokens into the key and value spaces, respectively. Based on these projections, {the enhanced visual representation $\textbf{v}_{\text{refined}}$} is formulated as:
\begin{align}
\label{eq:13}
&\textbf{v}_{\text{refined}} = \textbf{v}_\text{cls} + \nonumber \\
&\quad \text{Softmax}\left( \frac{(\textbf{v}_\text{cls}\mathbf{W}_Q^v)(\hat{\mathcal{T}}\mathbf{W}_K^v)^\top}{\sqrt{d}} \right) (\hat{\mathcal{T}}\mathbf{W}_V^v),
\end{align}
where $d$ is the token embedding dimension. Using Eq.~\ref{eq:13}, we selectively aggregate the semantics of local text keywords (\textit{e.g.}, specific attributes) into the global visual embeddings.

\noindent \textbf{Visual-Token to Textual.}
 Symmetrically, we use the global text token ${\mathbf{t}}_\text{eos}^i$ as the \textbf{query} to aggregate spatial details from the visual patches $\hat{\mathcal{V}}$ (serving as the \textbf{key} and \textbf{value}). Using another set of projection matrices $\{\mathbf{W}_Q^t, \mathbf{W}_K^t, \mathbf{W}_V^t\}$, { the enhanced textual representation $\textbf{t}_{\text{refined}}^i$ }is formulated as:
\begin{align}
\label{eq:14}
&\textbf{t}_{\text{refined}}^i = \textbf{t}_\text{eos}^i + \nonumber \\
&\quad \text{Softmax}\left( \frac{(\textbf{t}_\text{eos}^i\mathbf{W}_Q^t)(\hat{\mathcal{V}}\mathbf{W}_K^t)^\top}{\sqrt{d}} \right) (\hat{\mathcal{V}}\mathbf{W}_V^t)\,.
\end{align}
Using Eq.~\ref{eq:14} anchors the abstract class embedding to the current image instance by integrating specific  patch-level information, resulting in an enriched representation.\\
\textbf{Global Loss.} We use {the enhanced global visual representation $\textbf{v}_{\text{refined}}$ and textual representation $\textbf{t}_{\text{refined}}^i$ }to reformulate the prediction of class $i$ in Eq.~\ref{eq:clip_pred} and to calculate {the enhanced global contrastive loss $\mathcal{L}_{r}$}:
 \begin{align} \label{eq:unit-loss}
	\mathcal{L}_{r} &= \ell (f_\text{refined}(\mathbf{x}), y), \quad \text{where}\\ 
	\label{eq:unit-logit}
	f_{\text{refined},{y_i}}(\mathbf{x}) & =\frac{\exp(\cos(\textbf{v}_{\text{refined}}, \textbf{t}_{\text{refined}}^i)/\tau)}{\sum_{j} \exp(\cos(\textbf{v}_{\text{refined}}, \textbf{t}_{\text{refined}}^j)/\tau)}.
\end{align}
To further leverage the extracted local details, we also employ the cross-modal fusion mechanism on the local feature sets $ \hat{\mathcal{V}}$ and $\hat{\mathcal{T}}^i$. Specifically, to obtain the enhanced visual local feature set $ \hat{\mathcal{V}}_{\text{fused}}$, we replace the input $\textbf{v}_{\text{cls}}$ in Eq.~\ref{eq:13} with $ \hat{\mathcal{V}}$. Symmetrically, by replacing the input $\textbf{t}_\text{eos}^i$ in Eq.~\ref{eq:14}, we can obtain the enhanced textual local feature set $\hat{\mathcal{T}}^i_{\text{fused}}$. Subsequently, we take the average of these enhanced local feature sets to obtain {the aggregated local visual feature ${\textbf{v}_\text{local}}$ and textual feature ${\textbf{t}_\text{local}^i}$}:
  \begin{align}\label{eq:144}
      {\textbf{v}_\text{local}} = \text{AvgPool}(\hat{\mathcal{V}}_{\text{fused}}),\\\label{eq:15}
       {\textbf{t}_\text{local}^i} = \text{AvgPool}(\hat{\mathcal{T}}^i_{\text{fused}}).
  \end{align}
\textbf{Auxiliary Local Loss.} Similar to Eq.~\ref{eq:unit-loss}, we use {aggregated local features ${\textbf{v}_\text{local}}$ and ${\textbf{t}_\text{local}^i}$} to compute {the local contrastive loss $\mathcal{L}_{l}$}:
 \begin{align} \label{eq:local-loss}
	\mathcal{L}_{l} &= \ell (f_\text{local}(\mathbf{x}), y), \quad \text{where}\\ 
	\label{eq:local-logit}
	f_{\text{local},{y_i}}(\mathbf{x}) & =\frac{\exp(\cos(\textbf{v}_{\text{local}}, \textbf{t}_{\text{local}}^i) /\tau)}{\sum_{j} \exp(\cos(\textbf{v}_{\text{local}}, \textbf{t}_{\text{local}}^j) /\tau)}.
\end{align}
Finally, the loss function of the cross-modal fusion module $\mathcal{L}_{\text{cross}}$ can be expressed as a weighted combination of the global loss $\mathcal{L}_{r}$ and the auxiliary local loss $\mathcal{L}_{l}$:
\begin{align} 
\label{eq:cross-loss}
    \mathcal{L}_{\text{cross}} = \mathcal{L}_{r} +\lambda \mathcal{L}_{l},
\end{align}
where $\lambda$ is a hyperparameter that balances these two losses.

\noindent\textbf{Task-Specific Projector.} Despite the enhanced classification capability provided by the cross-modal fusion module, we observed that the model requires specific adaptation for various downstream tasks. To effectively adapt the model to downstream tasks while preserving its generalization ability, we introduce two \textit{task-specific} linear projectors, which consist of a task-specific visual projector $\textbf{M}_v(\cdot):\mathbb{R}^d\rightarrow\mathbb{R}^d$ and a textual projector $\textbf{M}_t(\cdot):\mathbb{R}^d\rightarrow\mathbb{R}^d$. These projectors are placed immediately after the top-$k$ local features selection for each incremental task $b$. To prevent catastrophic forgetting during  update, we instantiate new task-specific modules $\textbf{M}_v^b$ and $\textbf{M}_t^b$ during the training of task $b$. Simultaneously, we freeze the parameters of all historical modules $\{\textbf{M}_v^{1:b-1},\textbf{M}_t^{1:b-1}\}$ to preserve the learned representations of previous classes. Finally, we cumulatively aggregate the outputs from all frozen and active projectors. The projected visual feature $\tilde{\textbf{v}}_\text{cls} $  and textual feature $\tilde{\textbf{t}}_\text{eos}^i$ are computed as:
\begin{align}
\label{eq:12}
\tilde{\textbf{v}}_\text{cls} &= {\mathbf{M}}_v^b(\textbf{v}_\text{cls}) +\sum_{i=1}^{b-1}  \bar{\mathbf{M}}_v^i(\textbf{v}_\text{cls}),
\\
\label{eq:20}
        \tilde{\textbf{t}}_\text{eos}^i &= {\mathbf{M}}_t^b(\textbf{t}_\text{eos}^i) +\sum_{i=1}^{b-1} \bar{\mathbf{M}}_t^i(\textbf{t}_\text{eos}^i),
\end{align}
where $\bar{\mathbf{M}}_v^i(\cdot)$ and $\bar{\mathbf{M}}_t^i(\cdot)$ denote the frozen projectors corresponding to the visual and textual modalities in the $i$-th incremental task, respectively.
Eq.~\ref{eq:12} and Eq.~\ref{eq:20} ensure that the model remains stable and adapts efficiently to new tasks, without forgetting previously learned knowledge. So we use the cumulatively aggregated features (\textit{i.e.}, $\tilde{\textbf{v}}_{\text{cls}}, \tilde{\textbf{t}}_{\text{eos}}^i,\tilde{\mathcal{V}}$ and $\tilde{\mathcal{T}}^i$ ) from the task-specific projectors to replace 
the raw features (\textit{i.e.}, $\textbf{v}_{\text{cls}}, \textbf{t}_{\text{eos}}^i,\hat{\mathcal{V}}$ and $\hat{\mathcal{T}}^i$ ) originally fed into the cross-modal fusion module in Eq.~\ref{eq:13} and Eq.~\ref{eq:14}.\\
\textbf{Projected Loss.} We use the {projected visual feature $\tilde{\textbf{v}}_{\text{cls}}$ and textual feature $\tilde{\textbf{t}}_{\text{eos}}^i$} to compute the loss $\mathcal{L}_{\text{proj}}$:
\begin{align} \label{eq:loss}
	\mathcal{L}_{\text{proj}} &= \ell (f_\text{proj}(\mathbf{x}), y), \quad \text{where}\\ 
	\label{eq:logit}
	f_{\text{proj},{y_i}}(\mathbf{x}) & =\frac{\exp(\cos(\tilde{\textbf{v}}_\text{cls}, \tilde{\textbf{t}}_{\text{eos}}^i) / \tau)}{\sum_{j} \exp(\cos(\tilde{\textbf{v}}_\text{cls}, \tilde{\textbf{t}}_{\text{eos}}^j) /\tau)}.
\end{align}
To further mitigate catastrophic forgetting, following~\cite{zhu2021prototype}, we model the feature distribution for each past class $i$ using the extracted global visual features (in Eq.~\ref{eq:2}) as a Gaussian distribution $\mathcal{N}(\boldsymbol{\mu}_i, \boldsymbol{\Sigma}_i)$. Subsequently, we resample pseudo-features from this modeled distribution and incorporate them into the current task's global feature representation to enable effective feature-level replay of historical data.
\subsection{Local Knowledge Preservation}\label{sec:4.3}
Within the cross-modal fusion module, the weight matrices $\{\textbf{W}^Q$, $\textbf{W}^K$, $\textbf{W}^V\}$ are globally shared. Across incremental tasks, this shared structure is highly susceptible to catastrophic forgetting. Therefore, we employ a parameter update strategy based on subspace projection to preserve the critical local knowledge subspace of previous tasks.

After training task $b$, we capture the current weight $\textbf{W}^{\text{old}}$ and use SVD to approximate the critical knowledge subspace. The singular value decomposition of $\textbf{W}^{\text{old}}$ is $\textbf{W}^{\text{old}} = \mathbf{U} \boldsymbol{\Sigma} \mathbf{V}^\top,$ where $\mathbf{U}$ and $\mathbf{V}$ are orthogonal matrices, and $\boldsymbol{\Sigma} = \text{diag}(\sigma_1, \sigma_2, \ldots)$ contains the singular values. To identify the most informative subspace, we determine the rank $r$ that preserves a sufficient proportion of the cumulative energy carried by the squared singular values ($\sigma_i^2$). We choose the minimum rank $r$ such that the ratio of accumulated energy meets a predefined threshold $\gamma$:
\begin{align}
    r = \min\left\{r' \Big| \frac{\sum_{i=1}^{r'} \sigma_i^2}{\sum \sigma_i^2} \ge \gamma \right\}, \gamma\in (0,1].
\end{align}
To preserve the knowledge from previous tasks, we construct the orthogonal projection matrix $\mathbf{P}_{\text{old}}$ using the first $r$ left singular vectors $\mathbf{U}_{\text{old}}= [\mathbf{u}_1, \mathbf{u}_2, \cdots, \mathbf{u}_r]$ as follows:
\begin{align}\label{eq:25}
    \mathbf{P}_{\text{old}} = \mathbf{U}_{\text{old}} \mathbf{U}_{\text{old}}^\text{T},
\end{align}
where $\mathbf{P}_{\text{old}}$ defines the parameter directions that are sensitive to the performance of previous tasks.

During the adaptation to the incremental task $b'$, the weights are updated via gradient descent to $\textbf{W}^{\text{new}}$. To enforce preservation of old knowledge, we apply a projection correction to $\textbf{W}^{\text{new}}$ at the end of new task training. The final projected weight $\textbf{W}^{\text{proj}}$ is computed as a weighted combination of $\textbf{W}^{\text{new}}$'s projection onto the old knowledge subspace $ \mathbf{P}_{\text{old}}$ and its orthogonal complement space $\mathbf{I} - \mathbf{P}_{\text{old}}$:
\begin{align}\label{eq:proj}
    \textbf{W}^{\text{proj}} = \alpha (\mathbf{P}_{\text{old}} \textbf{W}^{\text{new}}) + (1 - \alpha) ((\mathbf{I} - \mathbf{P}_{\text{old}}) \textbf{W}^{\text{new}}),
\end{align}
where $\mathbf{I}$ is the identity matrix, and $\alpha$ is a hyperparameter. 
\subsection{Summary of \name} \label{sec:4.4}
In \mame, we  enhance CLIP's discriminative power  in CIL by effectively leveraging patch-token features. We extract top-$k$ discriminative local features. The cross-modal fusion module then explicitly injects this local knowledge into global representations to achieve precise alignment. Finally, a subspace projection strategy protects critical weights, mitigating catastrophic forgetting. During training, we jointly optimize the losses by combining Eq.~\ref{eq:cross-loss} and Eq.~\ref{eq:loss}:
\begin{align}\label{eq:total-loss}
\mathcal{L} = \mathcal{L}_{\text{proj}} + \mathcal{L}_{\text{cross}}.
\end{align}
\noindent\textbf{Inference.} After learning the $b$-th incremental task, given an input image $\mathbf{x}$, we first extract the visual token sequence using the frozen CLIP visual encoder and select the top-$k$ discriminative visual patches according to Eq.~\ref{eq:9}. For each seen class $i$, we encode its text prompt  with the frozen CLIP text encoder and obtain the corresponding top-$k$ discriminative textual tokens. The selected features are then transformed by the learned task-specific projectors to produce the aggregated representations $\tilde{\textbf{v}}_{\text{cls}}, \tilde{\textbf{t}}_{\text{eos}}^i,\tilde{\mathcal{V}}$ and $\tilde{\mathcal{T}}^i$ using Eq.~\ref{eq:12} and Eq.~\ref{eq:20}. Based on these projected features, we further use cross-modal fusion module to obtain the refined global representations $\textbf{v}_{\text{refined}}$
and $\textbf{t}_{\text{refined}}^i$, and the fused local representations $ {\textbf{v}_\text{local}} $ and ${\textbf{t}_\text{local}^i}$. Accordingly, the final prediction is obtained by aggregating the  scores computed in Eq.~\ref{eq:unit-logit}, Eq.~\ref{eq:local-logit}, and Eq.~\ref{eq:logit}:
\begin{align}
    f(\mathbf{x}) = f_{\text{refined}}(\mathbf{x}) + f_{\text{local}}(\mathbf{x}) + f_{\text{proj}}(\mathbf{x}).
\end{align}
\begin{algorithm}[t]

    \caption{{Training \name\ for CIL}}
    \label{alg:training}
    \raggedright
    {\bf Input}: Current task dataset $\mathcal{D}^b$; Current model: $f(\cdot)$; previous subspace projection matrix $\mathbf{P}_{\text{old}}$. \\
    {\bf Output}: Updated model;
    \begin{algorithmic}[1]
        \State Freeze all previous projectors $\{\mathbf{M}_v^{1:b-1}, \mathbf{M}_t^{1:b-1}\}$; \label{line:1} 
  \State  Initialize current projectors $\mathbf{M}_v^b, \mathbf{M}_t^b$;\label{line:2}
        \For{$(\mathbf{x}, y) \in \mathcal{D}^b$}  
            \State Extract visual features $\mathbf{V}$ via $\bar{g}_v(\mathbf{x})$;\label{line:4}
            \State Extract text features $\mathbf{T}^i$ for each class $i $ via $\bar{g}_t(\cdot)$;\label{line:5}
            \State Select top-$k$ visual patches and  top-$k$ text tokens;\label{line:6}
            \State Compute projected features via Eqs.~\ref{eq:12} and \ref{eq:20}; \label{line:7}
            \State Cross-modal fusion via Eqs.~\ref{eq:13},~\ref{eq:14},~\ref{eq:144} and \ref{eq:15}; \label{line:8}
            \State Compute $f_{\text{refined}}(\mathbf{x}), f_{\text{local}}(\mathbf{x}),$ and $ f_{\text{proj}}(\mathbf{x})$; \label{line:9}
            \State Compute the loss $ \mathcal{L}$ via Eq.~\ref{eq:total-loss}; update the model;\label{line:10}
        \EndFor
        \If{$b>1$} \label{line:12}
            \State Correct  each shared weight matrix via Eq.~\ref{eq:proj};
        \EndIf
        \State Update $\mathbf{P}_{\text{old}}$ from corrected weights via Eq.~\ref{eq:25};\label{line:15}
       
\noindent \Return the updated model;
    \end{algorithmic}
\end{algorithm}
\noindent\textbf{{Pseudo Code.}} {We give the pseudo-code of \name to illustrate
the training process in Algorithm~\ref{alg:training}. At each incremental stage $b$, the model takes the current task dataset $\mathcal{D}^b$ together with the previous subspace projection matrix $\mathbf{P}_{\text{old}}$ as input. Before optimization, all projectors learned from previous tasks are frozen (Line~\ref{line:1}), while new visual and textual projectors are initialized for the current task (Line~\ref{line:2}). During training, for each instance-label pair, we first extract the visual and textual features using the frozen CLIP encoders (Line~\ref{line:4} to Line~\ref{line:5}), and then apply the patch-token information selection module to select the top-$k$ discriminative visual patches and textual tokens (Line~\ref{line:6}). Based on these selected features, we compute the projected representations and perform cross-modal fusion to enhance both global and local features (Line~\ref{line:7} to Line~\ref{line:8}). The logits are then obtained for model optimization (Line~\ref{line:9}), followed by loss computation and parameter updating (Line~\ref{line:10}). After finishing the current task $b$, we further correct the shared fusion weights via subspace projection when $b>1$, and update the projection matrix $\mathbf{P}_{\text{old}}$ for the next incremental stage (Line~\ref{line:12} to Line~\ref{line:15}). The final updated model is then returned as the output of the training process.}

\begin{table*}[t]
	\vspace{-6mm}
	\caption{Average and last performance comparison of different methods {under B0 and B-half settings. \(\bar{\mathcal A}\) and \(\mathcal A_B\) denote the average and last performance, respectively.  }
		The best performance is shown in bold.  
		{\bf All methods are initialized with the same pre-trained CLIP for a fair comparison.} {Except for PROOF, all compared methods are evaluated without exemplars.} }\label{tab:benchmark}
	\vspace{0mm}
	\centering
	\resizebox{0.9\textwidth}{!}{%
		\begin{tabular}{@{}lccccccccccccccc}
			\toprule
			\multicolumn{1}{c}{\multirow{3}{*}{Method}}
			&
			\multicolumn{4}{c}{Aircraft }   & 
			\multicolumn{4}{c}{CIFAR100 }	&	
			\multicolumn{4}{c}{Cars }   
			\\ 
			& 
			\multicolumn{2}{c}{B0 Inc10}   & 
			\multicolumn{2}{c}{B50 Inc10}	&		
			\multicolumn{2}{c}{B0 Inc10}   & 
			\multicolumn{2}{c}{B50 Inc10}	& 
			\multicolumn{2}{c}{B0 Inc10}   & 
			\multicolumn{2}{c}{B50 Inc10}	& 
			\\  
			& 
			{$\bar{\mathcal{A}}$} & ${\mathcal{A}_B}$  
			& {$\bar{\mathcal{A}}$} & ${\mathcal{A}_B}$
			& {$\bar{\mathcal{A}}$} & ${\mathcal{A}_B}$ 
			&  {$\bar{\mathcal{A}}$} & ${\mathcal{A}_B}$  
			& {$\bar{\mathcal{A}}$} & ${\mathcal{A}_B}$
			& {$\bar{\mathcal{A}}$} & ${\mathcal{A}_B}$ 
			\\
			\midrule
			Finetune  &  3.16 & 0.96 & 1.72 & 1.05 & 7.84 & 4.44& 5.30 & 2.46& 3.14 & 1.10 & 1.54 & 1.13\\
			CoOp~\cite{zhou2022learning} & 14.54 & 7.14 & 13.05 & 7.77 & 47.00 & 24.24 & 41.23 & 24.12& 36.46 & 21.65& 37.40 & 20.87\\
			SimpleCIL~\cite{zhou2023revisiting} &59.24 & 48.09 & 53.05 & 48.09 & 84.15 & 76.63& 80.20 & 76.63& 92.04 & 86.85 & 88.96 & 86.85\\
			ZS-CLIP~\cite{radford2021learning} &26.66 & 17.22 & 21.70 & 17.22& 81.81 & 71.38& 76.49 & 71.38& 82.60 & 76.37& 78.32 & 76.37\\
			L2P~\cite{wang2022learning}  &47.19 & 28.29 &44.07&32.13& 82.74 & 73.03& 81.14 & 73.61& 76.63 & 61.82& 76.37 & 65.64 \\
			DualPrompt~\cite{wang2022dualprompt}  & 44.30& 25.83 &46.07&33.57 & 81.63 & 72.44& 80.12 & 72.57& 76.26 & 62.94& 76.88 & 67.55 \\
			CODA-Prompt~\cite{smith2023coda}  & 45.98 & 27.69 & 45.14 & 32.28& 82.43 & 73.43& 78.69 & 71.58& 80.21 & 66.47& 75.06 & 64.19 \\
			RAPF~\cite{huang2024class}   &  50.38  & 23.61 &  40.47 &  25.44 & 86.14 & 78.04 & 82.17 &  77.93  & 82.89 & 62.85 &  75.87 & 63.19\\
   CLG-CBM~\cite{Yu_2025_CVPR}  &  46.78  & 38.22 &51.99 &  42.45&  81.82 & 74.57  & 68.61 & 65.55 & 78.94 & 74.26    & 80.11  &73.62 \\
    {MG-CLIP~\cite{huang2025mind}}
& {48.33}
& {32.34}
& {26.28}
& {13.02}
& {74.95}
& {64.33}
& {31.59}
& {12.72}
& {89.74}
& {82.78}
& {85.62}
& {81.26} \\

{PROOF~\cite{zhou2023learning}}
& {64.35}
& {50.12}
& {52.17}
& {45.48}
& {84.65}
& {76.75}
& {80.16}
& {75.69}
& {90.74}
& {86.51}
& {88.00}
& {85.58} \\

{HERMAN~\cite{xie2025herman}}
& {55.54}
& {45.90}
& {50.21}
& {46.32}
& {84.75}
& {77.04}
& {82.77}
& {77.71}
& {88.30}
& {83.84}
& {85.22}
& {84.33} \\
\rowcolor{LightCyan}\name   & 65.26  &  50.74 &  55.79 &  49.05 &  86.92 & 79.22  &83.15 &  78.47 & 93.94 & 89.35 & 91.15 &  89.30\\
\rowcolor{LightCyan}
{\mame++}
& {\bf 67.31}
& {\bf 55.45}
& {\bf 58.43}
& {\bf 53.58}
& {\bf 87.53}
& {\bf 79.91}
& {\bf 83.52}
& {\bf 79.28}
& {\bf 94.44}
& {\bf 90.96}
& {\bf 92.25}
& {\bf 91.31}\\
		\end{tabular}
	}	
	\resizebox{0.9\textwidth}{!}{%
		\begin{tabular}{@{}lccccccccccccccc}
			\toprule
			\multicolumn{1}{c}{\multirow{3}{*}{Method}}
			& 
			\multicolumn{4}{c}{ImageNet-R }   & 
			\multicolumn{4}{c}{CUB }	&	\multicolumn{4}{c}{UCF }   
			\\ 
			& 
			\multicolumn{2}{c}{B0 Inc20}   & 
			\multicolumn{2}{c}{B100 Inc20}	&	\multicolumn{2}{c}{B0 Inc20}   & 
			\multicolumn{2}{c}{B100 Inc20}	& 
			\multicolumn{2}{c}{B0 Inc10}   & 
			\multicolumn{2}{c}{B50 Inc10}	& 
			\\  
			& 
			{$\bar{\mathcal{A}}$} & ${\mathcal{A}_B}$  
			& {$\bar{\mathcal{A}}$} & ${\mathcal{A}_B}$
			& {$\bar{\mathcal{A}}$} & ${\mathcal{A}_B}$ 
			&  {$\bar{\mathcal{A}}$} & ${\mathcal{A}_B}$  
			& {$\bar{\mathcal{A}}$} & ${\mathcal{A}_B}$
			& {$\bar{\mathcal{A}}$} & ${\mathcal{A}_B}$ 
			\\
			\midrule
			Finetune  & 1.37 & 0.43& 1.01 & 0.88& 2.06 & 0.64& 0.56 & 0.47& 4.51 & 1.59& 1.21 & 0.80\\
			CoOp~\cite{zhou2022learning} &60.73 & 37.52& 54.20 & 39.77& 27.61 & 8.57& 24.03 & 10.14& 47.85 & 33.46& 42.02 & 24.74\\
			SimpleCIL~\cite{zhou2023revisiting} & 81.06 & 74.48& 76.84 & 74.48& 83.81 & 77.52& 79.75 & 77.52& 90.44 & 85.68& 88.12 & 85.68\\
			ZS-CLIP~\cite{radford2021learning} &83.37 & 77.17& 79.57 & 77.17 & 74.38 & 63.06& 67.96 & 63.06& 75.50 & 67.64& 71.44 & 67.64\\
			L2P~\cite{wang2022learning}  &75.97 & 66.52 & 72.82 & 66.77&   70.87&57.93 & 75.64 &66.12 & 86.34 & 76.43& 83.95 & 76.62 \\
			DualPrompt~\cite{wang2022dualprompt}  &76.21 & 66.65 & 73.22 & 67.58&69.89 &57.46 & 74.40 &64.84 & 85.21 & 75.82& 84.31 & 76.35 \\
			CODA-Prompt~\cite{smith2023coda}  & 77.69 & 68.95 & 73.71 & 68.05& 73.12&62.98 &73.95&62.21 & 87.76 & 80.14&83.04 & 75.03 \\
			RAPF~\cite{huang2024class}  & 81.26  & 70.48 & 76.10 & 70.23 &  79.09 & 62.77& 72.82 & 62.93 & 92.28 & 80.33&90.31 & 81.55\\
			 CLG-CBM~\cite{Yu_2025_CVPR}  &79.19    & 72.03 & 75.20  & 70.50  & 78.11 & 72.14 &78.17  & 73.03   & 90.97 &88.45  & 92.32  & 89.12\\
        {MG-CLIP~\cite{huang2025mind}}
& {83.18}
& {77.20}
& {50.47}
& {35.37}
& {74.20}
& {64.25}
& {53.47}
& {34.78}
& {87.74}
& {80.83}
& {75.45}
& {59.42} \\

{PROOF~\cite{zhou2023learning}}
& {83.84}
& {78.40}
& {79.85}
& {77.27}
& {82.31}
& {76.64}
& {79.20}
& {76.37}
& {92.21}
& {87.65}
& {91.65}
& {88.90} \\

{HERMAN~\cite{xie2025herman}}
& {82.20}
& {78.25}
& {79.49}
& {77.73}
& {80.24}
& {74.43}
& {75.96}
& {73.11}
& {92.80}
& {88.86}
& {92.09}
& {88.67} \\
  \rowcolor{LightCyan}\name    & 85.03 & \bf79.38  & 80.92 & \bf79.23 & 85.94 & 79.90& 81.91 & 79.47 & 94.40 & 89.47 & 92.91 & 89.58 \\
\rowcolor{LightCyan}
{\mame++}
& {\bf 85.63}
& {79.08}
& {\bf 81.50}
& {78.02}
& {\bf 86.29}
& {\bf 80.92}
& {\bf 82.47}
& {\bf 79.77}
& {\bf 94.74}
& {\bf 90.45}
& {\bf 93.89}
& {\bf 91.10}\\
		\end{tabular}
	}
	
	\resizebox{0.9\textwidth}{!}{%
		\begin{tabular}{@{}lccccccccccccccc}
			\toprule
			\multicolumn{1}{c}{\multirow{3}{*}{Method}}
			&
			\multicolumn{4}{c}{SUN }   & 
			\multicolumn{4}{c}{Food }	&	\multicolumn{4}{c}{ObjectNet }   
			\\ 
			& 
			\multicolumn{2}{c}{B0 Inc30}   & 
			\multicolumn{2}{c}{B150 Inc30}	&		\multicolumn{2}{c}{B0 Inc10}   & 
			\multicolumn{2}{c}{B50 Inc10}	& 
			\multicolumn{2}{c}{B0 Inc20}   & 
			\multicolumn{2}{c}{B100 Inc20}	& 
			\\  
			& 
			{$\bar{\mathcal{A}}$} & ${\mathcal{A}_B}$  
			& {$\bar{\mathcal{A}}$} & ${\mathcal{A}_B}$
			& {$\bar{\mathcal{A}}$} & ${\mathcal{A}_B}$ 
			&  {$\bar{\mathcal{A}}$} & ${\mathcal{A}_B}$  
			& {$\bar{\mathcal{A}}$} & ${\mathcal{A}_B}$
			& {$\bar{\mathcal{A}}$} & ${\mathcal{A}_B}$ 
			\\
			\midrule
			Finetune  &4.51 & 1.59& 0.78 & 0.72& 3.49 & 1.71& 2.14 & 1.52& 1.34 & 0.47 & 0.69 & 0.54\\
			CoOp~\cite{zhou2022learning} &45.93 & 23.11 & 39.33 & 24.89& 36.01 & 14.18& 33.13 & 18.67& 21.24 & 6.29& 16.21 & 6.82\\
			SimpleCIL~\cite{zhou2023revisiting} & 82.13 & 75.58& 78.62 & 75.58& 87.89 & 81.65& 84.73 & 81.65& 52.06 & 40.13& 45.11 & 40.13\\
			ZS-CLIP~\cite{radford2021learning} &79.42 & 72.11& 74.95 & 72.11& 87.86 & 81.92& 84.75 & 81.92& 38.43 & 26.43 & 31.12 & 26.43\\	
			L2P~\cite{wang2022learning} &82.82 & 74.54 & 79.57 & 73.10 & 85.66 & 77.33& 80.42 & 73.13& 51.40 & 39.39& 48.91 & 42.83 \\
			DualPrompt~\cite{wang2022dualprompt} & 82.46 & 74.40 & 79.37 & 73.02& 84.92 &77.29& 80.00 & 72.75& 52.62 & 40.72& 49.08 & 42.92 \\
			CODA-Prompt~\cite{smith2023coda} &   83.34 & 75.71 & 80.38 & 74.17& 86.18 & 78.78& 80.98 & 74.13& 46.49 & 34.13& 40.57 & 34.13 \\
			RAPF~\cite{huang2024class}  & 82.13 & 72.47 & 78.04 & 73.10 & 88.57 & 81.15& 85.53 & 81.17&  48.67 & 27.43 & 39.28 &  28.73 \\
			 CLG-CBM~\cite{Yu_2025_CVPR}  & 84.04   & 76.19 &81.78   &76.39   &88.77  & 82.73 & 86.66 &  83.32  & 56.02 & 42.81 & 45.75  & 42.99\\ 
   {MG-CLIP~\cite{huang2025mind}}
& {53.71}
& {41.62}
& {26.58}
& {12.64}
& {88.59}
& {82.35}
& {28.86}
& {12.51}
& {51.41}
& {40.90}
& {25.00}
& {12.39} \\

{PROOF~\cite{zhou2023learning}}
& {81.10}
& {74.03}
& {80.15}
& {76.23}
& {87.12}
& {79.98}
& {82.85}
& {78.53}
& {56.07}
& {43.69}
& {48.90}
& {43.62} \\

{HERMAN~\cite{xie2025herman}}
& {83.17}
& {76.25}
& {79.92}
& {75.67}
& {88.30}
& {82.07}
& {85.94}
& {82.35}
& {55.18}
& {42.77}
& {48.95}
& {43.17} \\
\rowcolor{LightCyan}\name  & 84.15 & 76.90 & 80.87 & 76.86 & 88.81 & 82.99 &  86.80 & 83.74&  57.84  & 44.03  &  50.07&   43.90 \\
   \rowcolor{LightCyan}
{\mame++}
& {\bf 85.26}
& {\bf 78.46}
& {\bf 82.14}
& {\bf 78.84}
& {\bf 89.86}
& {\bf 83.77}
& {\bf 87.80}
& {\bf 84.95}
& {\bf 59.32}
& {\bf 45.38}
& {\bf 51.46}
& {\bf 44.98}\\
			\bottomrule
		\end{tabular}
	}
    \vspace{-3mm}
\end{table*}
\section{Experiments}

In this section, we conduct extensive experiments on nine benchmark datasets to evaluate the performance of \name and compare it to state-of-the-art methods. More experimental results are provided in the appendix.
\subsection{Implementation Details}
\noindent\textbf{Dataset.}  Following~\cite{zhou2023learning},~\cite{zhou2022learning}, we evaluate our method on nine benchmark datasets, \textit{i.e.}, {CIFAR100}~\cite{krizhevsky2009learning}, {CUB200}~\cite{WahCUB2002011}, {ObjectNet}~\cite{barbu2019objectnet}, {ImageNet-R}~\cite{hendrycks2021many}, {FGVCAircraft}~\cite{maji2013fine}, {StanfordCars}~\cite{krause20133d}, {Food101}~\cite{bossard2014food}, {SUN397}~\cite{xiao2010sun} and {UCF101}~\cite{soomro2012ucf101}. For data splitting, we adopt the sampling strategy of~\cite{zhou2023learning}, sampling 100 classes from CIFAR100, Aircraft, Cars, Food, UCF, 200 classes from CUB200, ObjectNet, ImageNet-R, and 300 classes from SUN.
\\\textbf{Dataset split.} Following~\cite{rebuffi2017icarl},~\cite{wang2022learning}, we utilize `B-$m$ Inc-$n$' to split the classes in CIL. Here, $m$ denotes the number of classes in the initial stage, while $n$ represents the number of classes in each subsequent incremental stage. For fair comparison, following~\cite{rebuffi2017icarl}, we randomly shuffle the class order using a random seed of 1993 and use the same order for all methods.
\\\textbf{Comparison methods.}
We first compare the performance against current SOTA CIL methods based on pre-trained models, \textit{e.g.}, L2P~\cite{wang2022learning}, DualPrompt~\cite{wang2022dualprompt}, CODA-Prompt~\cite{smith2023coda}, and SimpleCIL~\cite{zhou2023revisiting}. Furthermore, we compare with SOTA CLIP-based CIL methods, \textit{e.g.}, CoOp~\cite{zhou2022learning}, RAPF~\cite{huang2024class}, CLG-CBM~\cite{Yu_2025_CVPR},  {MG-CLIP~\cite{huang2025mind}, HERMAN~\cite{xie2025herman}, and PROOF~\cite{zhou2023learning}}. We use the finetune approach as a sequential fine-tuning baseline and adopt ZS-CLIP~\cite{radford2021learning} as a reference for CLIP on downstream tasks. \textbf{All compared methods are initialized using the same CLIP weights.}
\\ \textbf{Training details.} Our experiments are implemented in PyTorch 2.0.1 based on the C3Box~\cite{sun2026c3box} toolbox. To ensure a fair comparison, all compared methods are based on the CLIP with ViT-B/16 backbone, following~\cite{zhou2023learning},~\cite{huang2024class}. For vision-based methods that cannot leverage the textual prompt (\textit{e.g.}, L2P, DualPrompt, CODA-Prompt), we initialize them using CLIP's visual branch. Our method is optimized using SGD with a batch size of 64 over 10 epochs. The initial learning rate is set to 0.05 and is decayed via a cosine annealing scheduler. We set the energy threshold $\gamma$ for orthogonal projection to 0.9, the subspace weighting factor $\alpha$ to 0.9, the local feature selection size $k$ to 8, and the balance coefficient $\lambda$ to 0.15. Detailed analyses of these important hyperparameters are provided in Sec.~\ref{sec:5.3}. Our experimental results are based on the LAION-400M pre-trained CLIP model~\cite{ilharco2021openclip}.
\\\textbf{Evaluation metric.}
Following~\cite{rebuffi2017icarl},~\cite{zhou2023learning}, we let $\mathcal{A}_b$ represent the model's accuracy after the $b$-th incremental stage. We use two main metrics to evaluate our method: the last accuracy $\mathcal{A}_B$, which measures performance after the last task, and the average accuracy $\bar{\mathcal{A}}=\frac{1}{B}\sum_{b=1}^{B}\mathcal{A}_b$, which represents the mean accuracy across all stages.
\subsection{Benchmark Comparison} \label{sec:5.2}

\begin{figure*} 
\vspace{-5mm}
	\centering
	\begin{subfigure}{0.32\textwidth}
		\includegraphics[width=0.85\columnwidth]{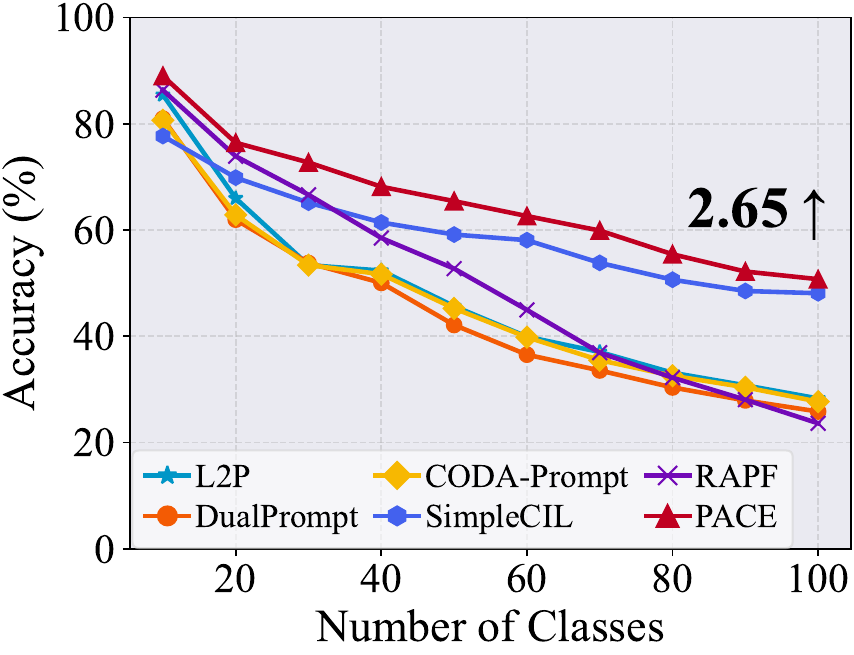}
		\caption{Aircraft Base0 Inc10}
	\end{subfigure}
	\hfill
	\begin{subfigure}{0.32\linewidth}
		\includegraphics[width=0.85\linewidth]{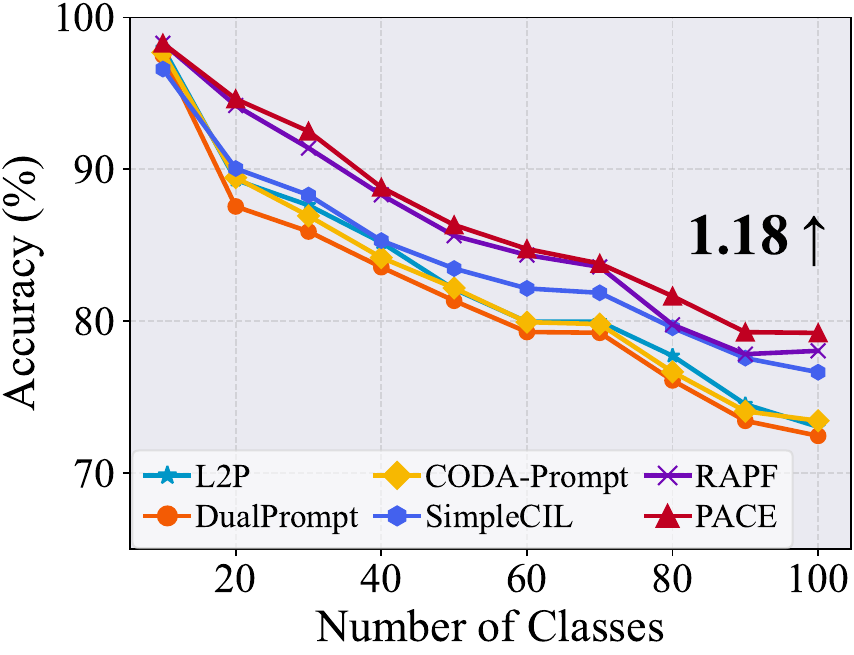}
		\caption{CIFAR100 Base0 Inc10}
	\end{subfigure}
	\hfill
	\begin{subfigure}{0.32\linewidth}
		\includegraphics[width=0.85\linewidth]{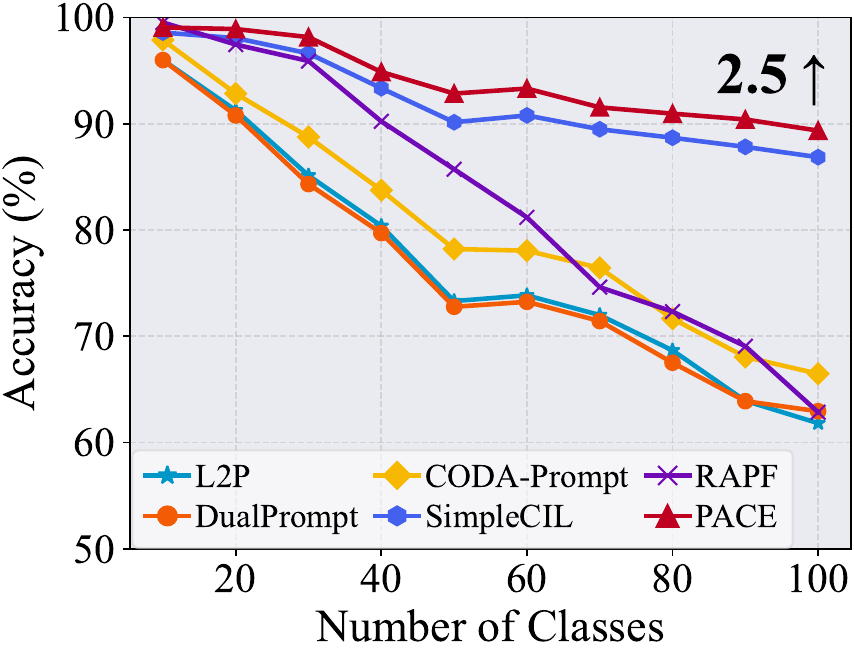}
		\caption{Cars Base0 Inc10}
	\end{subfigure}
	\\
	\vspace{-1mm}
	\caption{Incremental performance of different methods on the B0 setting. We report the performance gap after the last incremental stage of \name and the runner-up method at the end of the line.  \textbf{All methods utilize the same CLIP pre-trained weights.}}
		\vspace{-4mm}
	\label{fig:supp-benchmark-b0}
\end{figure*}
\begin{figure*}[ht]
  \vspace{-2mm}
 \centering
    \begin{subfigure}[t]{0.32\textwidth}
        \centering
        \includegraphics[width=0.85\linewidth]{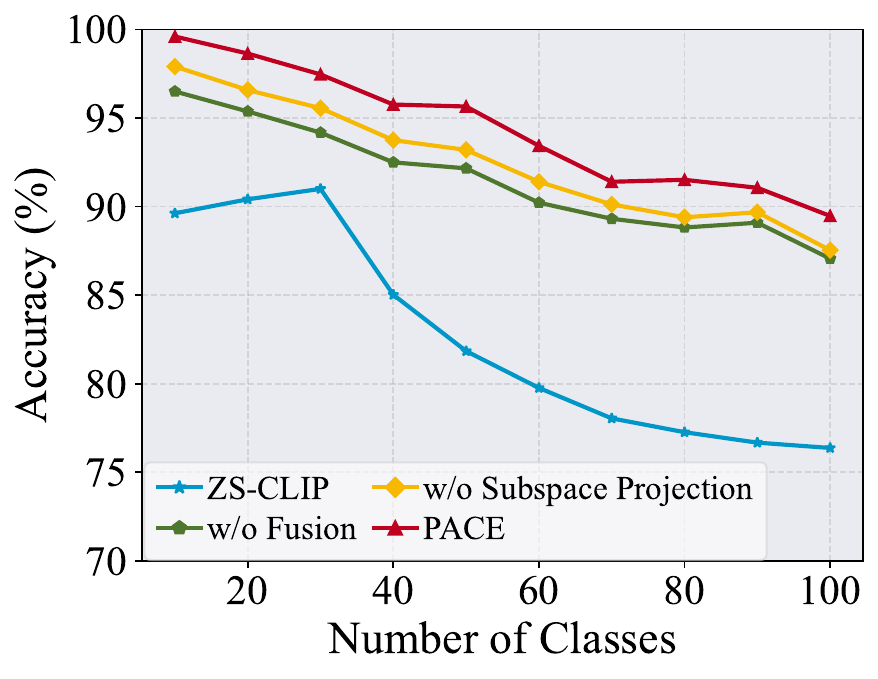}
        \caption{Ablation study}
        \label{fig:ablation}
    \end{subfigure}
    \hfill
   \begin{subfigure}[t]{0.32\textwidth}
        \centering
        \includegraphics[width=0.85\linewidth]{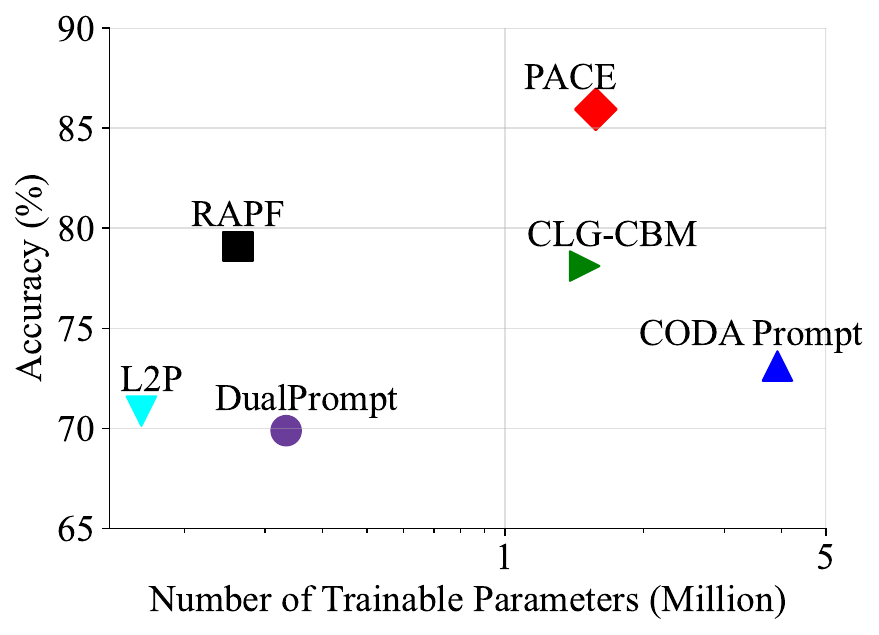}
        \caption{Trainable parameters}
        \label{fig:your-sixth-figure}
    \end{subfigure}
     \begin{subfigure}[t]{0.32\textwidth}
        \centering
        \includegraphics[width=0.85\linewidth]{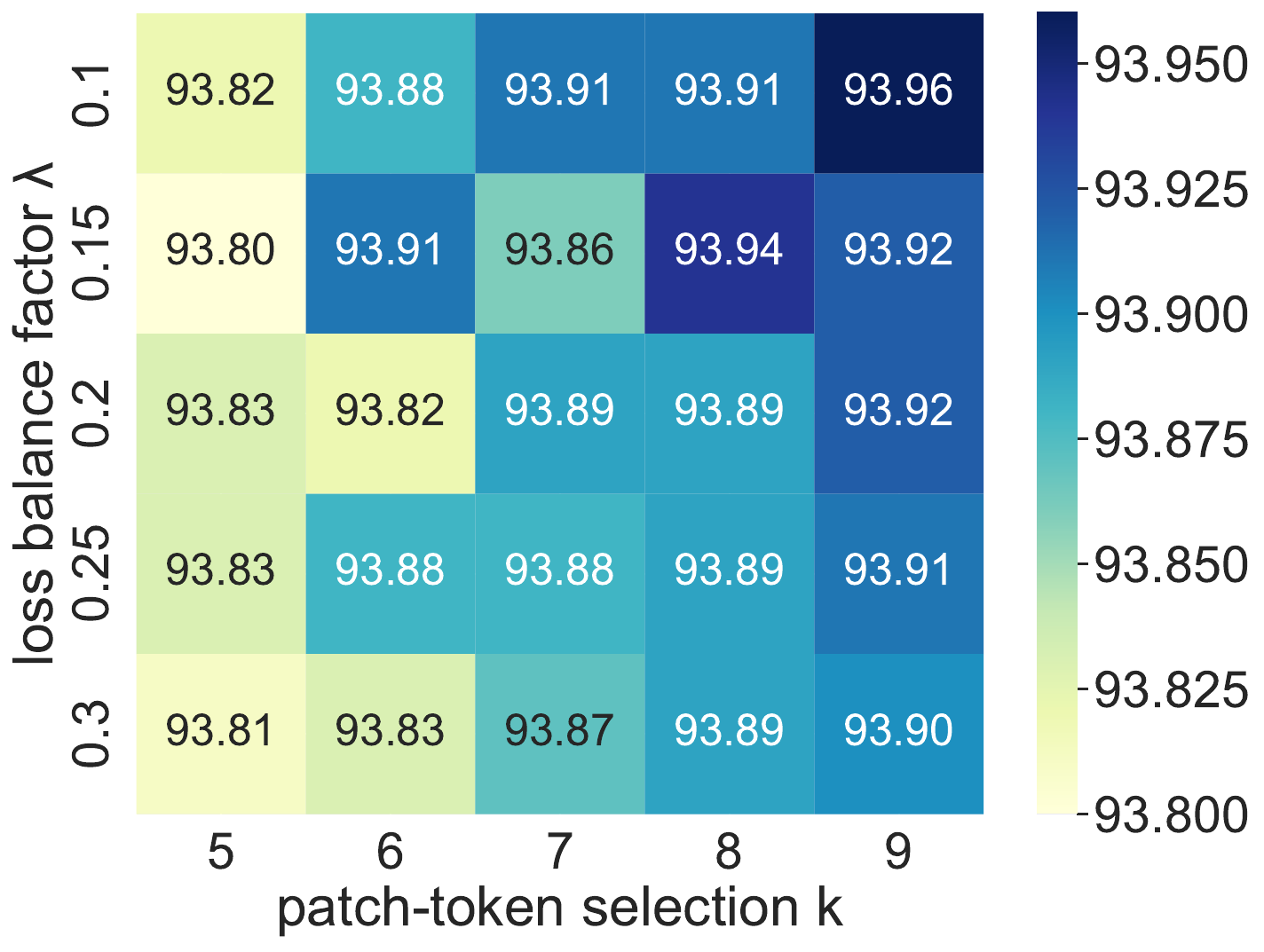}
        \caption{Parameter sensitivity}
        \label{fig:sensitivity}
    \end{subfigure}
    \vspace{-1mm}
    \caption{{{Ablation and sensitivity analyses of \mame. (a) Ablation study of the main components. (b) Comparison of trainable parameters and accuracy with representative CIL methods. (c) Sensitivity analysis of the patch-token selection number \(k\) and loss balance factor \(\lambda\)}}.}
    \vspace{-2mm}
    \label{fig:all-experiments}
\end{figure*}

We benchmark \name against SOTA methods on nine benchmark datasets and report the results in Tab.~\ref{tab:benchmark} and Fig.~\ref{fig:supp-benchmark-b0}. Specifically, \name consistently achieves superior performance. The Finetune baseline consistently performs the worst among all compared methods. Furthermore, we observe that visual prompt-based methods, such as L2P, DualPrompt, and CODA-Prompt, encounter performance ceilings primarily because they fail to leverage the rich textual semantics. Similarly, the textual prompt-tuning method CoOp suffers from prompt forgetting across tasks, resulting in degraded performance. Notably, \name outperforms other CLIP-based CIL methods, such as RAPF, further demonstrating its effectiveness in leveraging multimodal information and robustly mitigating catastrophic forgetting. This substantial advantage is attributed to our strategy of effectively leveraging critical localized information. Our cross-modal fusion module is specifically designed to achieve stronger cross-modal alignment by integrating information from visual patch tokens with token-level textual descriptors, thereby enabling a more robust and contextually aware representation for incremental tasks.

\subsection{Further Analysis}\label{sec:5.3}
\noindent\textbf{Ablation study.} We conduct detailed ablations on Cars B0 Inc10 to investigate the contribution of the core components of \mame, with results presented in Fig.~\ref{fig:ablation}. \name consistently maintains the highest accuracy across all incremental stages, demonstrating a widening performance gap over the baseline as the number of classes increases. Specifically, the \textbf{ZS-CLIP} baseline performs worst, and the removal of the cross-modal fusion denoted as \textbf{`w/o Fusion'} (\textit{i.e.}, not using Eq.~\ref{eq:unit-logit} and Eq.~\ref{eq:local-logit}), results in a substantial and sustained performance drop, confirming the necessity of explicitly fusing discriminative patch-level visual features and token-level textual semantics to maintain high classification power in CIL. {{Conversely, the removal of the local knowledge preservation mechanism, denoted as \textbf{`w/o Subspace Projection'} (\textit{i.e.}, removing Eq.~\ref{eq:proj}), shows a milder impact in early stages but incurs a rapid decline in later stages as interference from new tasks accumulates.}} This result highlights 
the vital role of the subspace projection strategy in stabilizing prior knowledge and effectively mitigating catastrophic forgetting.

\noindent\textbf{Trainable parameters.} We further report the detailed number of trainable parameters for each compared method in Fig.~\ref{fig:your-sixth-figure}. As shown in the figure, \name maintains a scale of trainable parameters comparable to other competitive methods. Specifically, we only update the task-specific parameters within the current task's projector, while keeping the parameters corresponding to all previous tasks frozen. Furthermore, the parameters of the cross-modal fusion module are globally shared throughout the entire continual learning sequence. This design allows \name to mitigate parameter expansion across sequential tasks while achieving superior performance. 

\begin{table}[t]
\vspace{-2mm}
\caption{{Sensitivity analysis of the energy threshold $\gamma$ on Aircraft B0 Inc10 and CIFAR100 B0 Inc10.}}
\label{tab:gamma}
\centering
\vspace{-1mm}
\resizebox{0.98\columnwidth}{!}{%

\begin{tabular}{@{}cccccc@{}}
\toprule
Dataset 
& $\gamma=0.75$ 
& $\gamma=0.80$ 
& $\gamma=0.85$ 
& $\gamma=0.90$ 
& $\gamma=0.95$ \\
\midrule
Aircraft
& 64.26 
& 64.67
& 64.84
& \bf 65.26 
& 64.40 \\

CIFAR100 
& 86.43 
& 86.74 
& 86.83
& \bf86.92
& 86.66 \\
\bottomrule
\end{tabular}
}
\vspace{-4mm}
\end{table}

\noindent\textbf{Parameter robustness.} We first evaluate the robustness of our method on Cars B0 Inc10 by varying two key parameters: the number of top-$k$ and the loss balance factor $\lambda$. Specifically, we set $k \in \{5,6,7,8,9\}$ and $\lambda \in \{0.1,0.15,0.2,0.25,0.3\}$. We report the last performance $\mathcal{A}_B$ in Fig.~\ref{fig:sensitivity}. As shown in the figure, the performance is robust with the change of these parameters. This indicates that the chosen $k$ already provides a stable balance between preserving discriminative local information and suppressing redundant noise, and we recommend $k =8$ and $\lambda = 0.15$ as the default values. {{We further conduct a sensitivity analysis to investigate the influence of the energy threshold \(\gamma\) on the final performance with \(\gamma \in \{0.75, 0.80, 0.85, 0.90, 0.95\}\) on Aircraft B0 Inc10 and CIFAR100 B0 Inc10. As shown in Table~\ref{tab:gamma}, the performance varies only slightly across different thresholds, while \(\gamma=0.9\) consistently achieves the best performance on both datasets.}}

\noindent\textbf{Patch selection strategies.} {To validate the effectiveness of the top-$k$ selection strategy, we further compare our method (Top-$k$) with Random-$k$ and Attention-only on Cars B0 Inc10 and ImageNet-R B0 Inc20. Random-$k$ randomly selects the same number of patches as Top-$k$, while Attention-only directly performs cross-modal enhancement using all patch tokens without explicit selection. The experimental results in Table \ref{tab:selection-ablation} show that the Top-$k$ strategy outperforms Random-$k$ and Attention-only, indicating that the proposed strategy more effectively preserves  discriminative local regions while suppressing redundant background noise.}
\begin{table}[t]
	\vspace{-3mm}
	\caption{Performance comparison under different top-$k$ patch selection strategies {on Cars B0 Inc10 and ImageNet-R B0 Inc20. \(\bar{\mathcal A}\) and \(\mathcal A_B\) denote the average and last performance, respectively. The best performance is shown in bold.}}
	\label{tab:selection-ablation}
	\centering
	\vspace{-1mm}
	\resizebox{0.98\columnwidth}{!}{%
		\begin{tabular}{@{}ccccc@{}}
			\toprule
			\multicolumn{1}{l}{\multirow{2}{*}{Patch Selection}} &
			\multicolumn{2}{c}{Cars B0 Inc10} &
			\multicolumn{2}{c}{ImageNet-R B0 Inc20} \\
			& {$\bar{\mathcal{A}}$} & {$\mathcal{A}_B$}
			& {$\bar{\mathcal{A}}$} & {$\mathcal{A}_B$} \\
			\midrule
			Random-$k$ & 89.33 & 82.48 & 81.16 & 72.79 \\
            {Attention-only} & {92.14} & {88.42} &{84.30} & {78.40} \\
              \rowcolor{LightCyan}
			Top-$k$ (ours)  & \bf93.94 & \bf89.35 &\bf85.03 & \bf79.38  \\
			\bottomrule
		\end{tabular}
	}
	\vspace{-0mm}
\end{table}

\noindent\textbf{{Computational Overhead Analysis.}} {We further compare the computational overhead of different CIL methods on Cars B0 Inc10. As shown in Table~\ref{tab:overhead_cars_b0_inc10}, \name achieves the highest average accuracy while maintaining moderate computational overhead. The patch-token information selection only requires affinity computation between global and patch-level features followed by lightweight top-$k$ selection. The cross-modal enhancement performs interactions only on selected tokens instead of the complete token sequences, while task-specific projectors introduce only lightweight linear projections during inference. Overall, \name achieves a favorable trade-off between recognition performance and computational efficiency.}

\begin{table}[t]
\vspace{-0mm}
    \caption{{Computational overhead comparison of representative CIL methods on Cars B0 Inc10.
  \textit{Eval time} denotes the evaluation time; \textit{Peak Memory} denotes the maximum training memory usage; \textit{Infer.} denotes the average inference latency over all test images and sessions; \textit{FLOPs} denotes the floating-point operations required for inference with a single image input. Higher is better for average accuracy, while lower values indicate lower computational overhead. The best results are highlighted in bold.}}
    \label{tab:overhead_cars_b0_inc10}
    \centering
    \resizebox{\columnwidth}{!}{%
    
   \begin{tabular}{@{}lccccc@{}}
\toprule
Method 
& $\bar{\mathcal{A}}\uparrow$
& Eval Time (min)$\downarrow$
& Peak Memory (GB)$\downarrow$
& Infer. (ms/img)$\downarrow$ & FLOPs (G)$\downarrow$\\
\midrule
ZS-CLIP~\cite{radford2021learning} &82.60& \textbf{0.96} & \textbf{1.04} &\textbf{2.56} &\textbf{11.29}\\
DualPrompt~\cite{wang2022dualprompt} &76.26&3.22&6.95&8.58 &28.96\\
CODA-Prompt~\cite{smith2023coda} &80.21&3.59&15.53&9.58&35.33\\
RAPF~\cite{huang2024class} &82.89&1.34&6.43&3.59&11.29\\
CLG-CBM~\cite{Yu_2025_CVPR} &78.94&1.13&{1.05}&3.01&11.29\\
PROOF~\cite{zhou2023learning} &90.74&2.41&1.16&6.43&11.85\\
\midrule
\rowcolor{LightCyan}
PACE (Ours)&\textbf{93.94}&\text{1.91}&1.74&\text{5.09}&12.41\\
\bottomrule
\end{tabular}
}
    \vspace{-5mm}
\end{table}

\begin{figure}[t]
\vspace{-3mm}
\centering

\begin{subfigure}[t]{0.48\columnwidth}
    \centering
    \includegraphics[width=0.95\linewidth]{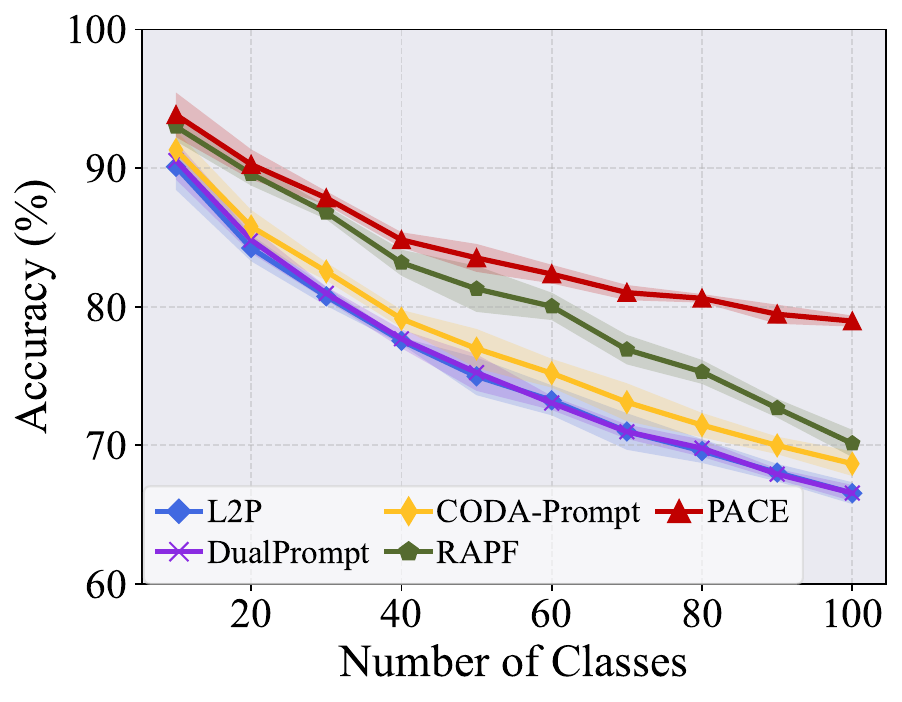}
    \vspace{-1mm}
    \caption{{Multiple runs}}
    \label{fig:seeds}
\end{subfigure}
\hfill
\begin{subfigure}[t]{0.48\columnwidth}
    \centering
    \includegraphics[width=0.95\linewidth]{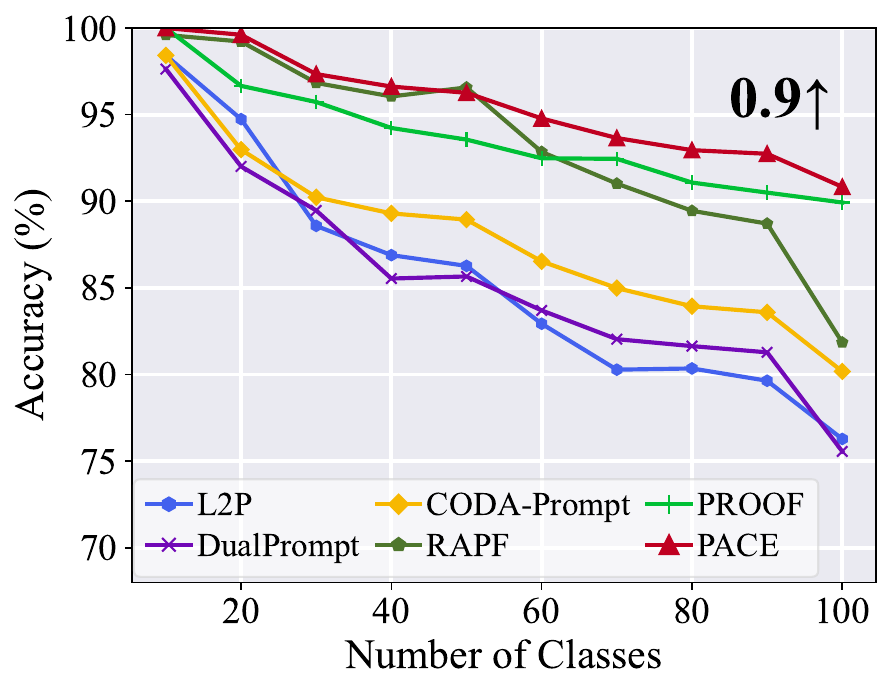}
    \vspace{-1mm}
    \caption{{OpenAI pre-trained weights}}
    \label{fig:backbone}
\end{subfigure}

\vspace{-1mm}
\caption{{Robustness analysis of \mame.
(a) Results on ImageNet-R B0 Inc20 under multiple random runs, where \name consistently achieves superior performance.
(b) Results on UCF B0 Inc10 using OpenAI pre-trained CLIP weights, where \name maintains superior performance.}}
\label{fig:robustness}
\vspace{-6mm}
\end{figure}

\noindent\textbf{{Multiple runs.}} {To investigate the robustness of different methods, we perform experiments across multiple runs with varying random seeds. Specifically, we generate five class splits using random seeds $\{1993, 1994, 1995, 1996, 1997\}$, and report the mean and standard deviation of the average and last performance on Aircraft B0 Inc10, Cars B0 Inc10, and ImageNet-R B0 Inc20. The corresponding results are summarized in Table~\ref{tab:2seed}. In addition, Fig.~\ref{fig:seeds} presents the performance variation across different runs on ImageNet-R B0 Inc20. The results show that \name consistently achieves superior performance across all experimental runs, highlighting its stability.}

\begin{table*}[t]
	\vspace{-5mm}
	\caption{{Robustness analysis under different random class orders. We report the average and last accuracy of different methods as \emph{mean}  $\pm$  \emph{std}  over multiple random seeds. The best results are highlighted in bold. \textbf{All methods are initialized from the same pre-trained CLIP backbone.}} 
	}\label{tab:2seed}
	\vspace{-2mm}
	\centering
	\resizebox{\textwidth}{!}{%
    
		\begin{tabular}{@{}lcccccccc}
			\toprule
			\multicolumn{1}{c}{\multirow{2}{*}{Method}}
			&
			\multicolumn{2}{c}{Aircraft B0 Inc10}   & 
			\multicolumn{2}{c}{Cars B0 Inc10}   
			&
			\multicolumn{2}{c}{ImageNet-R B0 Inc20}
			\\ 
			& 
			{$\bar{\mathcal{A}}$} & ${\mathcal{A}_B}$  
			& {$\bar{\mathcal{A}}$} & ${\mathcal{A}_B}$ 
			& {$\bar{\mathcal{A}}$} & ${\mathcal{A}_B}$
			\\
            \midrule
SimpleCIL~\cite{zhou2023revisiting} 
&61.05 $\pm$  {1.28} & 48.09 $\pm$  {0.00} 
& 91.54 $\pm$  {0.81} & 86.85 $\pm$  {0.00}
& 80.98 $\pm$  {0.57} & 74.48 $\pm$  {0.00}\\
			
  RAPF~\cite{huang2024class}   
&  45.81 $\pm$  {1.64}  & 23.78 $\pm$  {1.11} 
& 80.05 $\pm$  {1.34} & 69.70 $\pm$  {1.36}
& 82.93 $\pm$  {0.47} & 76.13 $\pm$  {0.58}\\
CLG-CBM~\cite{Yu_2025_CVPR}  
& 47.48 $\pm$  {1.57}   &38.60 $\pm$  {0.80} 
&78.66 $\pm$  {0.96}  & 74.45 $\pm$  {0.34} 
& 79.42 $\pm$  {0.24} & 72.00 $\pm$  {0.87}\\
  PROOF~\cite{zhou2023learning}  
&  66.34 $\pm$  {1.50}  & 50.82 $\pm$  {0.70} 
& 92.18 $\pm$ {0.97}  & 88.38 $\pm$  {0.17}
& 83.63 $\pm$  {0.46} & 78.71 $\pm$  {0.12}\\
 
  \rowcolor{LightCyan}\name 
&\textbf{67.55 $\pm$  {1.27}}    &\textbf{51.66 $\pm$  {0.58}} 
& \textbf{93.88 $\pm$  {0.91}} &  \textbf{89.63 $\pm$  {0.20}}
& \textbf{84.15 $\pm$  {0.57}} & \textbf{79.16 $\pm$  {0.24}}\\
			\bottomrule
		\end{tabular}
	}	
    \vspace{0mm}
\end{table*}

\begin{table*}[t]
	\vspace{-2mm}
	\caption{{Forgetting measure $F_B$ of different methods under different incremental settings. The best results are highlighted in bold. Lower $F_B$	indicates less forgetting.}}
	\label{tab:forgetting}
	\vspace{-2mm}
	\centering
	\resizebox{\textwidth}{!}{%
    
		\begin{tabular}{@{}lcccccccccccc}
			\toprule
			\multirow{2}{*}{Method} 
			& \multicolumn{2}{c}{Aircraft} 
			& \multicolumn{2}{c}{CIFAR100} 
			& \multicolumn{2}{c}{ImageNet-R} 
			& \multicolumn{2}{c}{SUN} \\
			& B0 Inc10 & B50 Inc10 
			& B0 Inc10 & B50 Inc10 
			& B0 Inc20 & B100 Inc20 
			& B0 Inc30 & B150 Inc30 \\
			\midrule 
            CODA-Prompt~\cite{smith2023coda} & 25.09& 20.03& 21.43& 19.48&16.60 &12.27 &22.71 & 17.69\\
			RAPF~\cite{huang2024class} & 25.68 &19.09 &\bf 8.14 & \bf 4.57& 9.95& \bf 4.48& 9.86& 4.81\\
		 CLG-CBM~\cite{Yu_2025_CVPR} &\bf 14.41 &\bf 5.32& 13.96&12.57 &11.44 &9.50 & 14.62& 12.26\\
		PROOF~\cite{zhou2023learning} &14.67 & 7.43&16.47 & 15.27& 9.06& 6.80& 15.16& 5.01\\
  \rowcolor{LightCyan}
		\name & 15.09 & 6.24 & 13.67& 8.45 &\bf 7.88 & 4.63 & \bf 8.78 & \bf 4.43 \\
			\bottomrule
		\end{tabular}
	}	
\vspace{-3mm}
\end{table*}

\noindent\textbf{{Different backbones.}}
 {In Sec.~\ref{sec:5.2}, we conduct the experiments by considering CLIP with ViT-B/16 under the LAION-400M~\cite{ilharco2021openclip} pre-trained weight. To extend our evaluation, we also report performance using OpenAI~\cite{radford2021learning} pre-trained weights. As illustrated in Fig.~\ref{fig:backbone} on UCF B0 Inc10, \name consistently yields superior performance across different pre-training backbones, affirming its effectiveness across different model initializations. }

\noindent\textbf{{Forgetting Measure.}} {To further analyze how much forgetting \name suffers on previously learned tasks, we additionally report the forgetting measure. Following~\cite{rebuffi2017icarl}, the forgetting measure is computed as the average performance drop from the best-achieved accuracy of each task to its final accuracy:
\begin{equation}
    F_B=\frac{1}{B-1}\sum_{b=1}^{B-1}\max_{l\in\{b,\dots,B-1\}}(\mathcal{A}_{l,b}-\mathcal{A}_{B,b}),
\end{equation}
where \(\mathcal{A}_{l,b}\) denotes the accuracy on task \(b\) after learning stage \(l\). We further evaluate the forgetting measure on Aircraft, CIFAR100, ImageNet-R, and SUN, and report the results in Table~\ref{tab:forgetting}. The results show that \name achieves consistently low forgetting measures across different CIL settings, demonstrating its effectiveness in preserving historical knowledge during incremental learning.}

\noindent\textbf{{Different  text prompt designs.}} {To analyze the impact of different text prompt designs on the performance of \mame, we provide an additional prompt sensitivity analysis. While keeping all other experimental settings and model components unchanged, we replace the original fixed prompt “A photo of a [CLASS]” with different text templates and compare the performance under different prompts. Specifically, besides the original template, we further evaluate several commonly used CLIP prompt templates, including “A photo of the [CLASS]”, “A good photo of a [CLASS]”, and “A blurry photo of a [CLASS]”. The results are summarized in Table~\ref{tab:different_prompt}. \name maintains stable performance under different prompt templates, indicating that the proposed method is not sensitive to a specific prompt and can effectively exploit textual semantic information from different prompt designs.
}
\begin{table}[t]
	\caption{{Prompt sensitivity analysis of \name under different text prompt templates on Cars B0 Inc10 and UCF B0 Inc10. The best performance is shown in bold.}}
	\label{tab:different_prompt}
	\centering
	\vspace{-1mm}
	\resizebox{0.98\columnwidth}{!}{%
    
		\begin{tabular}{@{}ccccc@{}}
			\toprule
			\multicolumn{1}{c}{\multirow{2}{*}{Prompt Template}} &
			\multicolumn{2}{c}{Cars B0 Inc10} &
			\multicolumn{2}{c}{UCF B0 Inc10} \\
			& {$\bar{\mathcal{A}}$} & {$\mathcal{A}_B$}
			& {$\bar{\mathcal{A}}$} & {$\mathcal{A}_B$} \\
			\midrule
			A photo of a [CLASS] & 93.94 & 89.35 & 94.40 & 89.47   \\
             
			A photo of the [CLASS]  & 93.92 & 89.30 & \bf 95.07 & \bf 90.00\\
            A good photo of a [CLASS] & \bf93.96 & \bf 89.44 & 94.84 & 90.22 \\
           A blurry photo of a [CLASS] & 93.91 & 89.27& 94.11 &  89.05 \\
			\bottomrule
		\end{tabular}
	}
	\vspace{-5mm}
\end{table}

\begin{figure*}
	\vspace{-0mm}
	\centering
	       \begin{minipage}{\textwidth}
        \centering
        \includegraphics[width=0.85\textwidth]{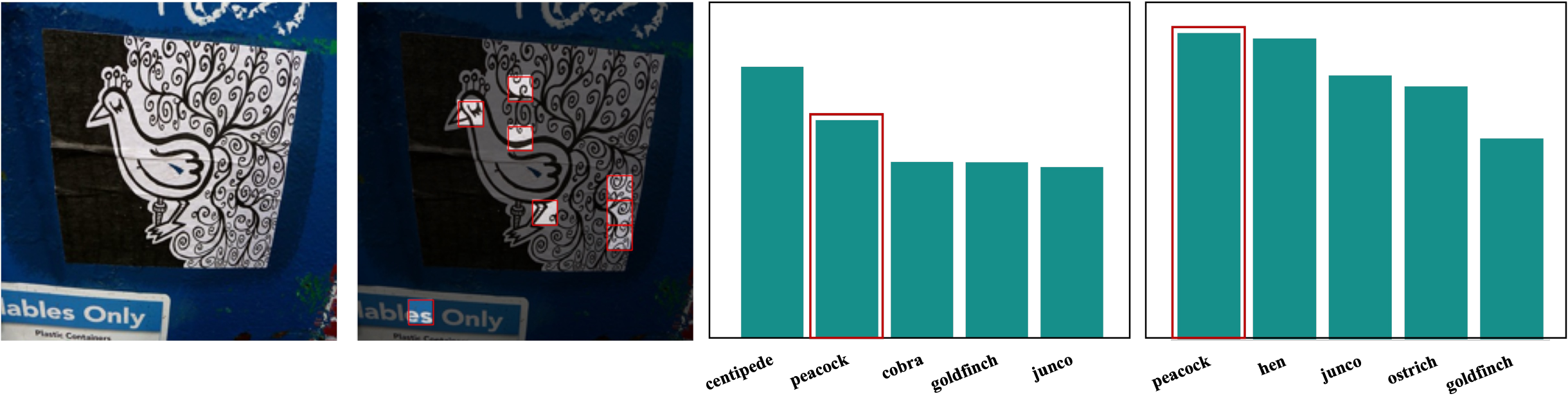} \\
        \label{fig:input}
    \end{minipage}
    
    \vspace{5mm}
    
    \begin{minipage}{\textwidth}
        \centering
        \includegraphics[width=0.85\textwidth]{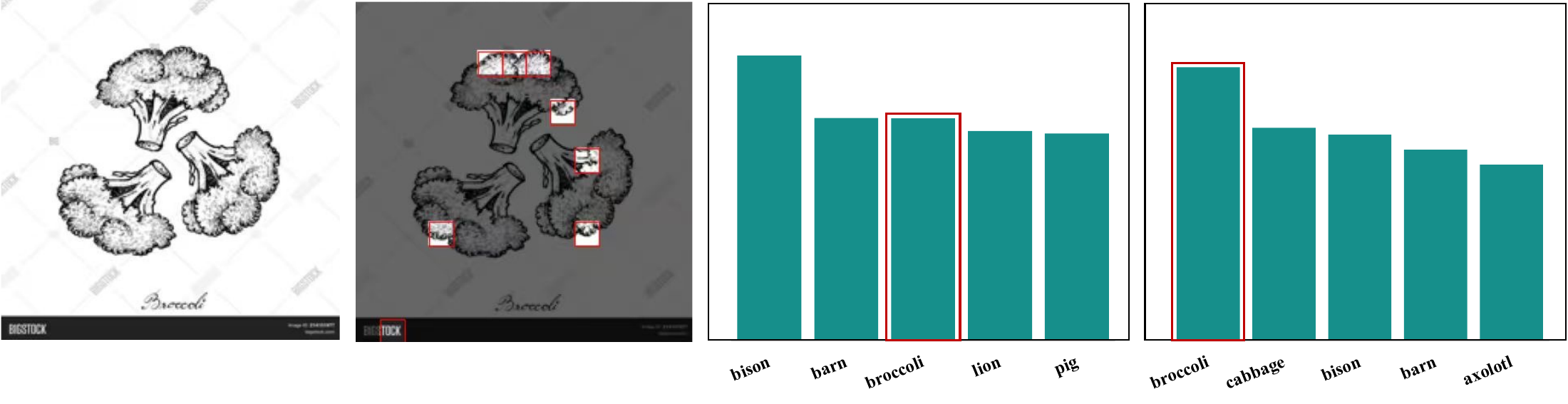} \\
        \label{fig:patches}
    \end{minipage}
 \vspace{-2mm}
	\caption{\small {\bf Column 1:} Input. {\bf Column 2:} Visualization of the top-$k$ discriminative patches selected. {\bf Column 3:} Top-5 predictions without the top-$k$ discriminative patches.  {\bf Column 4:} Top-5 predictions with the top-$k$ discriminative patches.}
	\vspace{-5mm}
	\label{fig:further-analysis}
\end{figure*}

\noindent\textbf{{Large-scale datasets.}} {To further evaluate the scalability and effectiveness of \name on large-scale datasets, we conduct additional experiments on ImageNet-100~\cite{deng2009imagenet} and ImageNet-1K~\cite{deng2009imagenet}. The experimental results are reported in Table~\ref{tab:datasets}. As shown in the table, \name consistently outperforms existing CIL methods on both large-scale benchmarks, demonstrating its effectiveness and scalability for large-scale datasets in CLIP-based CIL.}

\begin{table}[t]
	\caption{{Performance comparison on large-scale ImageNet-100 B0 Inc10 and ImageNet-1K B0 Inc100 benchmarks. The best performance is shown in bold.}}
	\label{tab:datasets}
	\centering
	\vspace{-1mm}
	\resizebox{0.98\columnwidth}{!}{%
    
		\begin{tabular}{@{}ccccc@{}}
			\toprule
			\multicolumn{1}{c}{\multirow{2}{*}{Method}} &
			\multicolumn{2}{c}{ImageNet-100 B0 Inc10} &
			\multicolumn{2}{c}{ImageNet-1K B0 Inc100} \\
			& {$\bar{\mathcal{A}}$} & {$\mathcal{A}_B$}
			& {$\bar{\mathcal{A}}$} & {$\mathcal{A}_B$} \\
			\midrule
			SimpleCIL~\cite{zhou2023revisiting}  & 91.10 & 86.88 & 75.94 & 68.37 \\
             
			RAPF~\cite{huang2024class}  & 91.92 & 87.02 & 73.09 & 65.59\\
           PROOF~\cite{zhou2023learning}  & 88.44 & 82.88 & 70.52  & 63.35\\
             \rowcolor{LightCyan}
           \name & \bf 92.21 & \bf 87.78 & \bf 77.43 & \bf 69.93\\
			\bottomrule
		\end{tabular}
	}
	\vspace{-6mm}
\end{table}

\noindent\textbf{Visualizations.}
We visualize the model output on ImageNet-R in Fig.~\ref{fig:further-analysis}. This figure is structured into four columns. Column 1 represents the original input image; Column 2 visualizes the extracted top-$k$ discriminative patch-tokens; Column 3 shows the top-5 predictions without using patch-token information; and Column 4 presents the top-5 predictions using patch-token information. We utilize bars with red edges to denote the ground-truth class.
As shown in the figure, using top-$k$ patch-token information effectively improves the representation of global features, thereby significantly enhancing the model's classification performance for the ground-truth class. Taking the broccoli sketch as an example, our method accurately captures the high-texture floret regions of paramount importance in the image. The effective introduction of this local information greatly strengthens the representation capability of the global feature, and directly and significantly enhances the final classification performance of the model for the ground-truth class ``Broccoli".

\noindent\textbf{{Fixed and adaptive patch selection.}} {To further examine the choice of \(k\) across different datasets, we evaluate \(k\in\{5,6,7,8,9\}\) on Aircraft B0 Inc10, Cars B0 Inc10, and ImageNet-R B0 Inc20. As shown in Table~\ref{tab:4topk}, \(k=8\) achieves the best performance on all three datasets. We further consider an adaptive token selection strategy. Specifically, we compute the cosine affinity between local tokens and their corresponding global representations, \textit{i.e.},
$\textbf{v}_\text{cls}$ and $\textbf{t}_\text{eos}^{i}$. The resulting scores are normalized to \([0,1]\) using min-max normalization. We then apply thresholds \(\delta\in\{0.7,0.8,0.9\}\) to dynamically determine the selected visual patches and textual tokens. The corresponding results are also reported in Table~\ref{tab:4topk}. Although \(\delta=0.8\) yields slight improvements on Aircraft and ImageNet-R, the gains are marginal, while the fixed \(k=8\) setting remains competitive and stable across datasets. Therefore, we adopt \(k=8\) as the default setting, as it provides stable and competitive performance across different datasets.}

\begin{table}[t]
\caption{{Comparison of fixed and adaptive token selection strategies  on Aircraft B0 Inc10, Cars B0 Inc10, and ImageNet-R B0 Inc20. The best results are highlighted in bold.}}
\label{tab:4topk}
\centering
\vspace{-1mm}
\resizebox{0.98\columnwidth}{!}{%

\begin{tabular}{@{}ccccccccc@{}}
\toprule
Dataset 
& $k=5$ 
& $k=6$ 
& $k=7$ 
& $k=8$ 
& $k=9$ 
&$\delta =0.7$
&$\delta =0.8$
&$\delta =0.9$
\\
\midrule
Aircraft
& 64.89 
& 64.90
&  65.12
&  65.26
& 64.84 
& 64.74 &  \bf 65.28 & 64.85\\

Cars 
& 93.80
& 93.91 
& 93.86
&\bf 93.94
& 93.92 
& 93.84 & 93.88 & 93.58\\

ImageNet-R  
& 84.50 
& 84.59
& 84.80
&  85.03
& 84.86  & 84.54 & \bf 85.07 & 84.38\\
\bottomrule
\end{tabular}
}
\vspace{-4mm}
\end{table}
\subsection{{Extending 
\name with LLM-Generated Descriptions}}
{To further investigate the capability of \name to exploit richer token-level textual semantics, we extend the original fixed prompt with LLM-generated class-specific visual descriptions and construct \mame++. Specifically, we employ GPT-5.4 to generate discriminative visual descriptions for each class. For a class \(y_i\), we provide its class name as the input to GPT-5.4~\cite{achiam2023gpt} with the following prompt:
\begin{displayquote}
\itshape
	 {\bf Q}: What are the key visual features for identifying a [CLASS]$_i$ in a photo? Focus on the most discriminative attributes.\\
	 {\bf A}: {\bf 1.} Pointed triangular ears. {\bf 2.} Vertical-slit pupils. {\bf 3.}$\cdots$
\end{displayquote}
Based on this instruction, GPT-5.4 generates a set of class-specific visual descriptions containing discriminative attributes, such as  shape, color, and local structures. These generated descriptions are then used to replace the original fixed prompt “A photo of a [CLASS]” and fed into the frozen CLIP text encoder to obtain the corresponding token-level textual features \(\mathbf{T}^{i}\).  Compared with \mame, \mame++ only modifies the textual prompt construction process by replacing the fixed template with LLM-generated visual descriptions. The frozen CLIP text encoder, subsequent token selection, and cross-modal enhancement module remain unchanged.  We further evaluate \mame++ on nine benchmark datasets to investigate the effectiveness of richer textual descriptions. As shown in Table~\ref{tab:benchmark}, \mame++ achieves  comparable overall performance and obtains improvements on several datasets compared with \mame, indicating that the generated visual descriptions can provide additional semantic information for local token alignment. }

\section{Conclusion}
Class-incremental learning is crucial for real-world applications. The \name method proposed in this paper aims to enhance CLIP's CIL capabilities by leveraging the rich patch-level visual features and token-level textual semantics that have largely been underutilized in CLIP. Specifically, we extract top-$k$ discriminative local features and subsequently use cross-modal fusion to leverage this local information to enhance global representations and alignment. Furthermore, a subspace-based knowledge preservation strategy effectively mitigates catastrophic forgetting in the shared fusion module. Extensive experiments validate \mame's superior performance.

\noindent\textbf{Limitations and Future Work.} Our current framework utilizes a fixed prompt for the textual branch, which may limit the model's ability to capture semantic diversity and fully exploit the power of the vision-language model in specific CIL scenarios. In the future, we will explore a richer prompt engineering strategy to further enhance textual representation and cross-modal alignment capability, \textit{e.g.}, learnable prompts.

\setcounter{section}{0}
\renewcommand{\thesection}{\Roman{section}}
\makeatletter

\bibliographystyle{IEEEtran}

\bibliography{reference}

\end{document}